\documentclass[aoas,MSNbibl,nameyear,dvips]{arximspdf}
\usepackage{mathbh}
\usepackage{graphicx}

\doi{10.1214/14-AOAS717} 
\volume{8}
\issue{2}
\pubyear{2014}
\firstpage{1145}
\lastpage{1181}

\makeatletter
\newtheorem{theorem}{Theorem}
\newtheorem{theoremstar}{Theorem}
\newtheorem{proposition}[theorem]{Proposition}
\makeatother

\begin{document}
\begin{frontmatter}

\title{Bayesian nonparametric Plackett--Luce models for the analysis
of preferences for\\ college degree programmes}
\runtitle{\hspace*{-5pt}BNP Plackett--Luce models for the analysis of preferences}

\begin{aug}
\author[a]{\fnms{Fran\c cois} \snm{Caron}\corref{}\ead[label=e1]{francois.caron@stats.ox.ac.uk}\thanksref{m1}\thanksref{t1}},
\author[a]{\fnms{Yee Whye} \snm{Teh}\ead[label=e2]{y.w.teh@stats.ox.ac.uk}\thanksref{m1}}\\
\and
\author[b]{\fnms{Thomas Brendan} \snm{Murphy}\ead[label=e3]{brendan.murphy@ucd.ie}\thanksref{m2}}
\runauthor{F. Caron, Y. W. Teh and T. B. Murphy}
\affiliation{University of Oxford\thanksmark{m1} and University
College Dublin\thanksmark{m2}}
\address[a]{F. Caron\\
Y. W. Teh\\
Department of Statistics\\
University of Oxford\\
1 South Parks Road\\
Oxford, OX1 3TG\\
United Kingdom\\
\printead{e1}\\
\phantom{E-mail:\ }\printead*{e2}} 
\address[b]{T. B. Murphy\\
School of Mathematical Sciences\\
University College Dublin\\
Dublin 4\\
Ireland\\
\printead{e3}}
\end{aug}
\thankstext{t1}{Supported by the European Commission under the Marie Curie Intra-European Fellowship Programme.}

\received{\smonth{1} \syear{2013}}
\revised{\smonth{1} \syear{2014}}

%
\begin{abstract}
In this paper we propose a Bayesian nonparametric model for clustering
partial ranking data.
We start by developing a Bayesian nonparametric extension of the
popular Plackett--Luce choice model that can handle an infinite number
of choice items. Our framework is based on the theory of random atomic
measures, with the prior specified by a completely random measure. We
characterise the posterior distribution given data, and derive a simple
and effective Gibbs sampler for posterior simulation. We then develop a
Dirichlet process mixture extension of our model and apply it to
investigate the clustering of preferences for college degree programmes
amongst Irish secondary school graduates. The existence of clusters of
applicants who have similar preferences for degree programmes is
established and we determine that subject matter and geographical
location of the third level institution characterise these clusters.
\end{abstract}

\begin{keyword}
\kwd{Ranking data}
\kwd{permutations}
\kwd{gamma process}
\kwd{Dirichlet process}
\kwd{mixture models}
\end{keyword}

\end{frontmatter}

\setcounter{footnote}{1}

\section{Introduction}\label{sec1}
In this paper we consider partial ranking data consisting of ordered
lists of the top-$m$ items among a set of objects. Data in the form of
partial rankings arise in many contexts.
For example, in this paper we shall consider data pertaining to the top
ten preferences of Irish secondary school graduates who are applying to
undergraduate degree programmes offered in Irish third level
institutions. The third level institutions consist of universities,
institutes of technologies and private colleges. This application is
described in detail in Section~\ref{collegeapplications}.

The Plackett--Luce model [\citet{Luce1959}; \citet{Plackett1975}] is
a popular model for modeling such partial rankings of a finite
collection of $M$ items.
It has found many applications,
including choice modeling [\citet{Luce1977}; \citet{Chapman1982}], sport
ranking [\citet{Hunter2004}] and voting [\citet{Gormley2008a}].
\citet{Diaconis1988}, Chapter~9, provides detailed discussions
on the
statistical foundations of this model.

In the Plackett--Luce model,
each item $k\in[M]=\{1,\ldots,M\}$ is assigned a positive rating
parameter $w_{k}$, which represents the desirability or rating of a
product in the case of choice modeling,
or the skill of a player in sport rankings.
The Plackett--Luce model assumes the following generative story for a
top-$m$ list
$\rho=(\rho_{1},\ldots,\rho_{m})$ of items $\rho_i\in[M]$:
at each stage $i=1,\ldots,m$, an item is chosen to be the $i$th
item in the list from among the items that have not yet
been chosen, with the probability that $\rho_i$ is selected being
proportional to its desirability $w_{\rho_i}$.
The overall probability of a given partial ranking $\rho$
is then
%
\begin{equation}
P(\rho)=\prod_{i=1}^{m}\frac{w_{\rho_{i}}}{ ( \sum_{k=1}^{M}%
w_{k} ) - ( \sum_{j=1}^{i-1}w_{\rho_{j}} ) }
\label{eqfinitepl}
\end{equation}
with the denominator in (\ref{eqfinitepl}) being the sum over all
items not yet selected at stage~$i$.

In many situations the collection of available items can be very large
and/or potentially unknown.
In this case a nonparametric approach can be sensible,
where the pool of items is assumed to be infinite and the model allows
for the possibility of items not
observed in previous top-$m$ lists to appear in future ones. A~na\"ive
approach, building upon recent work on Bayesian inference for the
(finite) Plackett--Luce model and its extensions [\citet{Gormley2009};
\citet{Guiver2009}; \citet{Caron2012}], is to first derive a Markov chain Monte Carlo
sampler for the finite model, then to ``take the infinite limit'' of
the sampler, where the number of available items becomes infinite, but
such that all unobserved items are grouped together for computational
tractability.

Such an approach, outlined in Section~\ref{model}, is reminiscent of a
number of previous approaches deriving the (Gibbs sampler for the)
Dirichlet process mixture model as the infinite limit of (a Gibbs
sampler for) finite mixture models [\citet{Nea1992b}; \citet{Ras2000a};
\citet{IshZar2002a}]. Although
intuitively appealing, this is not a satisfying approach since it is
not clear what the underlying nonparametric model actually is, as it is
actually the algorithm whose infinite limit was taken. It also does not
directly lead to more general and flexible nonparametric models with no
obvious finite counterpart, nor does it lead to alternative
perspectives and characterisations of the same model, or resultant
alternative inference algorithms. \citet{Orbanz2009} further
investigates the approach of constructing nonparametric Bayesian models
from finite-dimensional parametric Bayesian models.

\citet{Caron2012a} recently proposed a Bayesian
nonparametric\break  Plackett--Luce model based on a natural representation of
items along with their ratings as an atomic measure. Specifically, the
model assumes the existence of an infinite pool of items $\{X_k\}
_{k=1}^\infty$, each with its own rating parameter, $\{w_k\}
_{k=1}^\infty$. The atomic measure then consists of an atom located at
each $X_k$ with a mass of $w_k$:
%
\begin{equation}
G = \sum_{k=1}^\infty w_k
\delta_{X_k}. \label{eqG}
\end{equation}
The probability of a top-$m$ list of items, say, $(X_{\rho_1},\ldots,X_{\rho_m})$, is then a direct extension of the finite case (\ref
{eqfinitepl}):
%
\begin{equation}
P(X_{\rho_1},\ldots,X_{\rho_m}|G)=\prod
_{i=1}^{m}\frac{w_{\rho
_{i}}}{ ( \sum_{k=1}^{\infty}%
w_{k} ) - ( \sum_{j=1}^{i-1}w_{\rho_{j}} ) }. \label
{eqinfinitepl}
\end{equation}
Using this representation, note that the top item $X_{\rho_1}$ in the
list is simply a draw from the probability measure obtained by
normalising $G$, while subsequent items in the top-$m$ list are draws
from probability measures obtained by first removing from $G$ the atoms
corresponding to previously picked items and normalising. Described
this way, it is clear that the Plackett--Luce model is none other than a
partial size-biased permutation of the atoms in $G$ [\citet{PatTai1977a}], and the existing machinery of random measures and
exchangeable random partitions [\citet{Pitman2006}; \citet{Lijoi2010}] can be brought to bear on our problem.

For example, we may use a variety of existing stochastic processes to
specify a prior over the atomic measure $G$. \citet{Caron2012a} considered the case, described in Section~\ref{crmpl},
where $G$ is a gamma process. This is a completely random measure
[\citet{Kin1967a}; \citet{Lijoi2010}]
with gamma
marginals, such that the corresponding normalised probability measure
is a Dirichlet
process [\citet{Fer1973a}]. They showed that with the
introduction of a suitable set of auxiliary variables, it is possible
to characterise the posterior law of $G$ given observations of top-$m$
lists distributed according to (\ref{eqinfinitepl}). A simple Gibbs
sampler can then be derived to simulate from the posterior distribution
which corresponds to the infinite limit of the Gibbs sampler for finite
models. In the \hyperref[sec8]{Appendix}, we show that the construction can be extended
from gamma processes to general completely random measures, and we
discuss extensions of the Gibbs sampler to this more general case.

In Section~\ref{secmixture} we describe a Dirichlet process mixture
model [\citet{Fer1973a}; \citet{Lo1984a}] for heterogeneous
partial ranking data, where each mixture component is a gamma process
nonparametric Plackett--Luce model. As shown in Section~\ref
{collegeapplications}, such a model is relevant for capturing
heterogeneity in preferences for college degree programmes. As we will
see, in this model it is important to allow the same atoms to appear
across the different random measures of the mixture components,
otherwise the model becomes degenerate with all observed items that
ever appeared together in some partial ranking being assigned to the
same mixture component. To allow for this, we use a tree-structured
extension of the time-varying model of \citet{Caron2012a}. In Section~\ref{seccao} we apply this mixture model to
the Irish college degree programme preferences data, showing that the
model is able to recover clusters of students with similar and
interpretable preferences.

Finally, we conclude in Section~\ref{secdiscussion} with a discussion
of the important contributions of this paper and proposals for future work.

\section{Irish college degree programmes}\label{sec2}
\label{collegeapplications}

Applications to college degree programmes in Ireland are handled by a
centralised applications system called the College Application Office
(CAO) (\surl{www.cao.ie}); a degree programme involves studying a
specific subject (broad or focussed) in a particular third level
institution. The CAO handles applications for 35 different third level
institutions including universities, institutes of technologies and
private colleges. In the autumn of each year, a list of all degree
programmes for the subsequent year is made available to applicants.
Quite often new degree programmes are added to the list of potential
choices after the initial list has been published, thus meaning that
the potential list of degree programme choices is evolving and not
always completely known. Applications are completed early in the year
in which the students plan to enter their college degree programme. The
list of available degree programmes changes from year to year but has
been generally growing in size year on year. Many degree programmes
have a specific subject area, for example, Mathematics, History or
Computer Science, but others are more general, for example, Science,
Commerce or Arts. In the year 2000, which we are examining herein,
there were 533 degree programmes available to be selected by the
applicants. When students apply for degree programmes they rank up to
ten degree programmes, in order of preference, from the list of all
degree programmes that are being offered. Two examples of such
applications for two different applicants are shown in Table~\ref
{tabexdatacao}.

%
\begin{table}
\tabcolsep=0pt
\caption{Two samples from the CAO preference data. Each rank
observation is an ordered list of up to ten degree programmes}\label{tabexdatacao}
\begin{tabular*}{\tablewidth}{@{\extracolsep{\fill}}@{}lccc@{}}
\hline
\textbf{Rank} & \textbf{CAO code} & \textbf{College} & \textbf{Degree programme} \\
\hline
\phantom{0}1&DN002 & University College Dublin & Medicine\\
\phantom{0}2&GY501 & NUI-Galway & Medicine\\
\phantom{0}3&CK701 & University College Cork & Medicine\\
\phantom{0}4&DN006 & University College Dublin & Physiotherapy\\
\phantom{0}5&TR053 & Trinity College Dublin & Physiotherapy\\
\phantom{0}6&DN004 & University College Dublin & Radiotherapy\\
\phantom{0}7&TR007 & Trinity College Dublin & Clinical speech\\
\phantom{0}8&FT223 & Dublin IT & Human nutrition\\
\phantom{0}9&TR084 & Trinity College Dublin & Social work\\
10&DN007 & University College Dublin & Social science\\[3pt]
\phantom{0}1 &MI005 & Mary Immaculate Limerick & Education-primary teaching\\
\phantom{0}2 & CK301 & University College Cork & Law\\
\phantom{0}3 & CK105 & University College Cork & European studies\\
\phantom{0}4 & CK107 & University College Cork & Language-french\\
\phantom{0}5 & CK101 & University College Cork & Arts\\
\hline
\end{tabular*}     \vspace*{-3pt}
\end{table}

Places in these degree programmes are allocated on the basis of the
applicants' performance in the Irish Leaving Certificate examination.
Students typically take between seven and nine subjects in the Leaving
Certificate examination. Points between zero and one hundred are
awarded for each applicant's best six subjects in the Leaving
Certificate examination and the points are totalled to give an overall
points score. The allocation of applicants to most degree programmes is
solely on the basis of the applicant's points score and applicants with
a high points score are more likely to get their high preference
choices. The minimum points score of all applicants accepted into a
degree programme is publicly available and is called the points
requirement. It is worth mentioning that even though degree programmes
may have\vadjust{\goodbreak} required Leaving Certificate subjects and grades as part of
the minimum entry requirements, the subjects used in the applicant's
points score calculation can be any six Leaving Certificate subjects.

The college applications system in Ireland is much debated in the
educational sector and it receives much attention in the Irish media.
The debate has two main parts: one part of the debate is whether the
current system of allocating points to students on the basis of a
single Leaving Certificate examination is a fair method, especially
when the points can be gained from any Leaving Certificate subjects;
the other part of the debate explores the choice behaviour of the
applicants and whether students are choosing degree programmes in a
coherent manner. We focus on the applicant's choices which are core to
the second part of the debate.

Many people feel that students do not necessarily pick degree programmes
on the basis of the courses offered but that they choose on other
grounds, like the perceived prestige of the degree programme. However,
other factors like geographical location of the third level institution
may also have an impact on the applicant's choice behaviour. The two
example applications in Table~\ref{tabexdatacao} illustrate that a
number of factors influence applicants choices. The first applicant has
selected degree programmes in medicine and other health sciences, so
their choices appear to be largely based on the course material.
However, the second application includes a wide variety of different
degree programmes; the applicant's first choice degree programme leads
to a career in Primary Teaching, whereas the other degree programmes
are in different areas. However, the institutions that have been chosen
are geographically close (within 100~km).\vadjust{\goodbreak}

In the year 1997, the Department of Education and Science commissioned
a review of the Irish college applications system. A report [\citet{Hyland1999}] reviewed the current system and made some
recommendations concerning the future of the system. In addition, four
research reports were published, one of which [\citet{Tuohy1998}]
examined the applicant's choices. \citet{Tuohy1998} used a number
of exploratory data analysis techniques to investigate the degree
programmes selected, but without reference to the preference ordering,
and he found that subject matter was an important factor in applicant
choices. More recently, \citet{Gormley2006} used a
finite mixture of Plackett--Luce models to find clusters of applications
with similar choice profiles. They fitted their model using maximum
likelihood and chose the number of mixture components using the
Bayesian \mbox{Information} Criterion (BIC). Their results also indicated that
subject matter and geographical location were strong determinants of
student choices. However, the model fitting paradigm used in their
analysis could not find small clusters of applicants because of the
manner that BIC penalises each additional mixture component. Further,
\citet{McNicholas2007} used association rule mining to
further explore college applicant choices, but he restricted his
attention to degree programme choice combinations that were selected by
at least 0.5\% of the applicants; thus, that analysis emphasised only
high frequency choice behaviour.

\citet{OConnell2006} conducted a survey of
new college entrants (as opposed to applicants) in 2004 and found that
the choice of college where they commenced their degree programme was
influenced primarily by reputation and geographical location of the
third level institution, and that the choice of degree programme was
influenced by intrinsic interest in the subject matter and, to a lesser
extent, future career prospects. Whilst that study only looks at
students who entered college and the degree programme that they
ultimately studied, it provides a further insight into the factors that
influence choice of degree programme.

We investigate the complete degree programme choice data for the year
2000 cohort of applications to the College Application Office; these
data correspond to top-10 rankings of college degree programmes for
53,757 applicants. The model proposed herein has a number of appealing
properties because it can account for choosing from the large number of
degree programmes on offer, it allows for small differences in
preference between degree programmes, it facilitates discovering large
and small clusters of applicants with similar preferences, and the
fitting in the Bayesian paradigm facilitates a deep exploration of the
clustering and co-clustering of applicants.

\section{An extension of the Plackett--Luce model to countably infinite choice sets}\label{sec3}\label{model}

We start this section with a review of a Bayesian approach to inference
in finite Plackett--Luce models [\citet{Gormley2009};
\citet{Guiver2009}; \citet{Caron2012}]
and take the infinite limit\vadjust{\goodbreak} to arrive at a nonparametric model. This
will give good intuitions for how the model operates, before we
rederive the same nonparametric model more formally in the next section
using gamma processes.

Recall that we have $M$ choice items indexed by $[M]=\{1,\ldots,M\}$,
with item $k\in[M]$ having a positive desirability parameter $w_k$.
We will suppose that our data consists of $L$ partial rankings of the
$M$ choice items, with the $\ell$th ranking being denoted $\rho_{\ell
}=(\rho_{\ell1},\ldots,\rho_{\ell m})$, for $\ell=1,\ldots,L$,
where each $\rho_{\ell i}\in[M]$. For notational simplicity we assume
that all the partial rankings are of length $m$.

\subsection{Finite Plackett--Luce model with gamma prior}\label{sec3.1}

As noted in the\break  \hyperref[sec1]{Introduction}, the Plackett--Luce model constructs a
partial ranking $\rho_{\ell}=(\rho_{\ell1},\ldots,\rho_{\ell m})$
iteratively. At the $i$th stage, with $i=1,2,\ldots,m$, we pick $\rho
_{\ell i}$ as the $i$th item from among those not yet picked with
probability proportional to $w_{\rho_{\ell i}}$. The probability of
the partial ranking $\rho_\ell$ is then as given in (\ref
{eqfinitepl}). An alternative Thurstonian interpretation, which will
be important in the following, is as follows: for each item $k$ let
$z_{\ell k}$ be exponentially distributed with rate $w_k$:
\[
z_{\ell k}\sim\operatorname{Exp}(w_k).
\]
Thinking of $z_{\ell k}$ as the arrival time of item $k$ in a race, let
$\rho_{\ell i}$ be the index of the $i$th item to arrive [the index of
the $i$th smallest value among $(z_{\ell k})_{k=1}^M$]. The resulting
probability of the first $m$ items to arrive being $\rho_\ell$ can be
shown to be the\vadjust{\goodbreak} probability~(\ref{eqfinitepl}) from before. In this
interpretation $(z_{\ell k})$ can be understood as latent variables,
and the EM algorithm
[\citet{DemLaiRub1977a}] can be applied to
derive an algorithm to find a ML setting for the parameters
$(w_k)_{k=1}^M$ given\vspace*{1pt} multiple partial rankings. Unfortunately the
posterior distribution of $(z_{\ell k})_{k=1}^M$ given~$\rho_\ell$ is
difficult to compute, so we can instead consider an alternative
parameterisation: let $Z_{\ell i}$ be the waiting time for the $i$th
item to arrive after the $i-1$th item. That is,
\[
Z_{\ell i}=z_{\rho_{\ell i}}-z_{\rho_{\ell  i-1}}
\]
with $z_{\rho_{\ell0}}$ defined to be 0. Then it is easily seen that
the joint probability of the observed partial rankings, along with the
alternative latent variables $(Z_{\ell i})$, is
%
\begin{eqnarray} \label{eqfinitejoint}
&& P\bigl((\rho_\ell)_{\ell=1}^L,\bigl((Z_{\ell i})_{i=1}^{m}
\bigr)_{\ell
=1}^L|(w_k)_{k=1}^M
\bigr)
\nonumber\\[-8pt]\\[-8pt]
&&\qquad  = \prod_{\ell=1}^L \prod
_{i=1}^m w_{\rho_{\ell
i}}\exp \Biggl(-Z_{\ell i}
\Biggl( \sum_{k=1}^M w_k -
\sum_{j=1}^{i-1} w_{\rho_{\ell j}} \Biggr)
\Biggr).\nonumber
\end{eqnarray}
In particular, the posterior of $(Z_{\ell i})_{i=1}^m$ is simply
factorised, with
\[
Z_{\ell i}|(\rho_\ell)_{\ell=1}^L,(w_k)_{k=1}^M
\sim\operatorname {Exp} \Biggl(\sum_{k=1}^M
w_k - \sum_{j=1}^{i-1}
w_{\rho_{\ell
j}} \Biggr)
\]
being exponentially distributed.
The M step of the EM algorithm can be easily derived as well. The
resulting algorithm was first proposed by \citet{Hunter2004} as
an instance of the MM (majorisation--maximisation) algorithm [\citet{Lange2000}] and its reinterpretation as an EM
algorithm was recently given by \citet{Caron2012}.

Taking a further step, we note that the joint probability (\ref
{eqfinitejoint}) is conjugate to a factorised gamma prior over the
parameters, say, $w_k\sim\operatorname{Gamma}(\frac{\alpha}{M},\tau
)$ with hyperparameters $\alpha,\tau>0$. Now Bayesian inference can
be carried out, for example, using a variational Bayesian EM algorithm
or a Gibbs sampler. In this paper we shall consider only Gibbs sampling
algorithms. By regrouping the terms in the exponential in (\ref
{eqfinitejoint}), the parameter updates are derived to be [\citet{Caron2012}]:
%
\begin{eqnarray}
w_k|\rho,(Z_{\ell i}),(w_{k'})_{k'\neq k} &
\sim&\operatorname {Gamma} \Biggl( \frac{\alpha}{M}+n_k, \tau+\sum
_{\ell=1}^L\sum_{i=1}^m
\delta_{\ell ik} Z_{\ell i} \Biggr), \label{eqfiniteposterior}
\end{eqnarray}
where $n_k$ is the number of occurrences of item $k$ among the observed
partial rankings and
\[
\delta_{\ell ik}=\cases{ 0,&\quad if there is a $j< i$ with $
\rho_{\ell j}=k$,
\cr
1,&\quad otherwise.}
\]

Note that the definitions of $n_k$ and $\delta_{\ell ik}$ slightly
differ from those in \citet{Hunter2004} and \citet{Caron2012}. In these articles, the authors consider full
$m$-rankings of subsets of $[M]$, whereas we consider here partial
top-$m$ rankings of all $M$ items.

\subsection{Taking the infinite limit}\label{sec3.2}

A Gibbs sampler for a nonparametric\break  Plackett--Luce model can now be
easily derived by taking the limit as the number of choice items
$M\rightarrow\infty$. If item $k$ has appeared among the observed
partial rankings, the limiting conditional distribution (\ref
{eqfiniteposterior}) is well defined since $n_k>0$. For items that did
not appear in the observations, (\ref{eqfiniteposterior}) becomes
degenerate at 0. Instead we can define $w_*=\sum_{k\dvtx n_k=0} w_k$ to be
the total desirability among all the infinitely many unobserved items.
Making use of the fact that sums of independent gammas with the same
scale parameter is a gamma with shape parameter given by the sum of the
shape parameters,
\begin{eqnarray*}
w_*|\rho,(Z_{\ell i}),(w_{k})_{k\dvtx n_k>0} &\sim&
\operatorname {Gamma} \Biggl( \alpha, \tau+\sum_{\ell=1}^L
\sum_{i=1}^m Z_{\ell i} \Biggr).
\end{eqnarray*}
The resulting Gibbs sampler alternates between updating the latent
variables $(Z_{\ell i})$ and updating the desirabilities of the
observed items $(w_k)_{k\dvtx n_k>0}$ and of the unobserved ones $w_*$.

This nonparametric model allows us to estimate the probability of
seeing new items appearing in future partial rankings in a coherent
manner. While intuitive,\vadjust{\goodbreak} the derivation is ad hoc, in the sense that it
arises as the infinite limit of the Gibbs sampler for finite
Plackett--Luce models, and is unsatisfying, as it did not directly
capture the structure of the underlying infinite-dimensional object,
which we will show in the next section to be a gamma process.

\section{A Bayesian nonparametric Plackett--Luce model based on the gamma process}\label{sec4}\label{crmpl}

Let $\mathbb{X}$ be a measurable space of choice items. In the case of
college applications, the space $\mathbb{X}$ is the space of all
possible Irish programme courses. A~gamma process is a completely
random measure over $\mathbb{X}$ with gamma marginals. Specifically,
it is a random atomic measure of the form (\ref{eqG}), such that for
each measurable subset $A$, the (random) mass $G(A)$ is gamma
distributed. Assuming that $G$ has no fixed atoms [i.e., for each
element $x\in\mathbb{X}$ we have $G(\{x\})=0$ with probability one]
and that the atom locations $\{X_k\}$ are independent of their masses
$\{w_k\}$ (i.e., the gamma process is homogeneous), it can be shown
that such a random measure can be constructed as follows [\citet{Kin1967a}, Chapter~9]: each $X_k$ is i.i.d. according to a base
distribution $H$ [which we assume is nonatomic with density $h(x)$],
while the set of masses $\{w_k\}$ is distributed according to a Poisson
process over $\mathbb{R}^+$ with mean intensity
\[
\lambda(w) = \alpha w^{-1}e^{-w\tau},
\]
where $\alpha>0$ is the concentration parameter and $\tau>0$ the inverse
scale. We write this as $G\sim\Gamma(\alpha,\tau,H)$. Under this
parametrisation, we have that $G(A)\sim\operatorname{Gamma}
(\alpha H(A),\tau)$. $\lambda(w)h(x)$ is known as the L\'evy
intensity of the homogeneous CRM $G$. The jump part $\lambda(w)$ of
the L\'evy intensity verifies the necessary condition
%
\begin{equation}
\int_0^{\infty} \bigl(1-\exp(-w)\bigr)\lambda(w)\,dw
< \infty
\end{equation}
and plays a significant role in characterising the properties of the
gamma process.\vadjust{\goodbreak}

\begin{figure}

\includegraphics{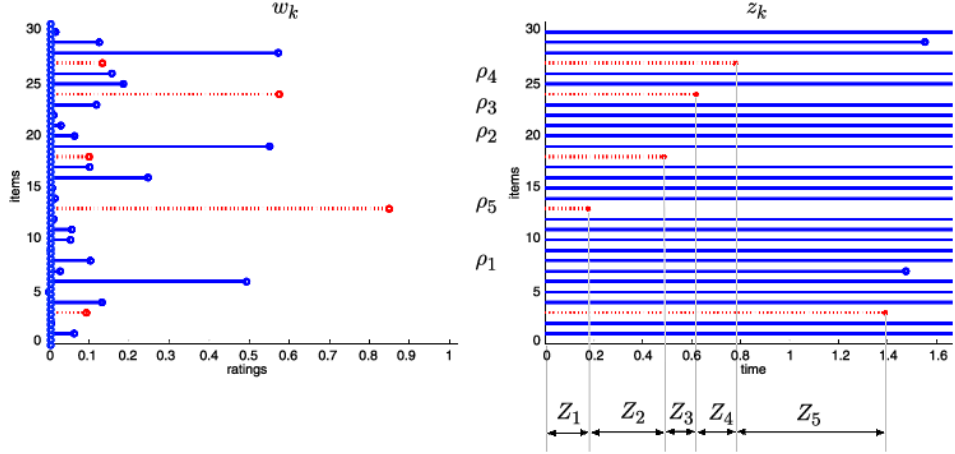}

\caption{Bayesian nonparametric Plackett--Luce model.
Left: an instantiation of the atomic measure $G$ encapsulating both the
items and their ratings. Right: arrival times $z_k$ and latent
variables $Z_k=z_{\rho_k}-z_{\rho_{k-1}}$. The top 5 items are $(\rho
_1,\rho_2,\ldots,\rho_5)$.}%
\label{figCRM}%
\end{figure}

\begin{figure}[b]

\includegraphics{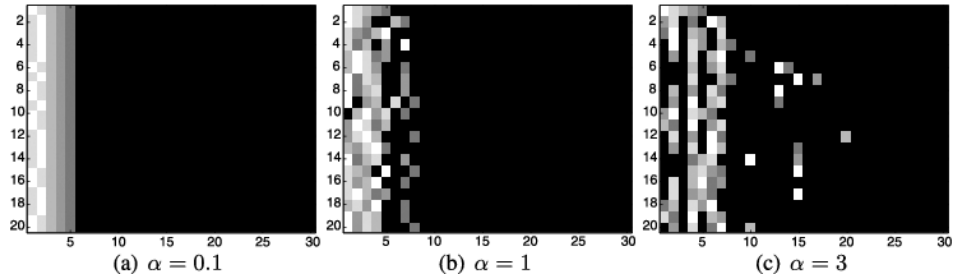}

\caption{Visualisation of top-5 rankings with rows corresponding to
different rankings and columns to items sorted by size-biased order. A
lighter shade corresponds to a higher rank. Results are shown for a
gamma process with $\lambda(w)=\alpha w^{-1}\exp(-\tau w)$ with $\tau
=1$ and different values of $\alpha$. The parameter $\alpha$ tunes
the variability in the partial rankings. The larger $\alpha$, the
higher the variability. As the probability of partial rankings \protect
\ref{eqinfinitepl} is invariant to rescaling of the weights, the
scaling parameter $\tau$ has no effect on the partial rankings.}\vspace*{-3pt}\label{figdraws}
\end{figure}

\begin{figure}

\includegraphics{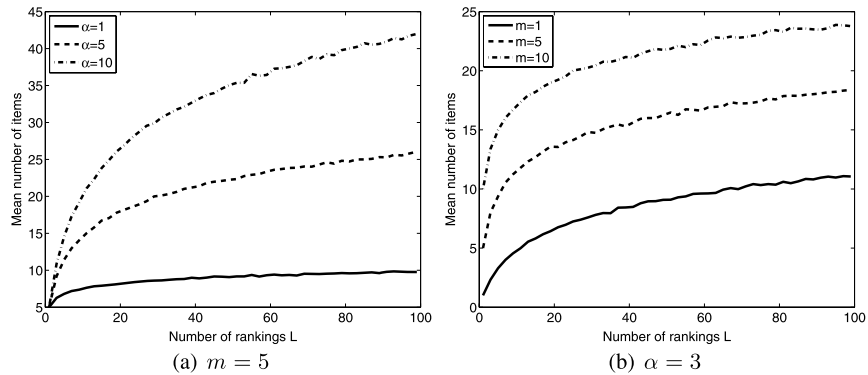}

\caption{Mean number of items appearing in $L$ top-$m$ rankings for a
gamma process with $\lambda(w)=\alpha w^{-1}\exp(-\tau w)$ with $\tau
=1$ and different values of $\alpha$ and $m$.}\vspace*{-3pt}\label{figmeanitems}
\end{figure}

We shall interpret each atom $X_k$ as a choice item, with its mass $w_{k}>0$
corresponding to the desirability parameter. The
Thurstonian view described in the finite model can be easily extended
to the nonparametric one, where a partial ranking $(X_
{\rho_1}, \ldots, X_{\rho_m})$ can be generated as the first $m$ items
to arrive in a race. In particular, for each atom $X_k$ let $z_k\sim
\operatorname{Exp}(w_k)$ be the time of arrival of
$X_k$ and $X_{\rho_i}$ the $i$th item to arrive. The first $m$
items to arrive $(X_{\rho_1}, \ldots, X_{\rho_m})$ then constitute
our partial ranking, with probability as given in (\ref
{eqinfinitepl}). This construction is depicted in Figure~\ref
{figCRM}. The top row of Figure~\ref{figdraws} visualises some top-5
rankings generated from the model, with $\tau=1$ and different values
of $\alpha$. Figure~\ref{figmeanitems} shows the mean number of
items appearing in $L$ top-$m$ rankings. For $m=1$, one recovers the
well-known result on the number of clusters for a Dirichlet process
model.

%

%

%

Again reparametrising using inter-arrival durations, let $Z_{i}=z_{\rho
_{i}}-z_{\rho_{i-1}}$ for $i=1,2,\ldots$ (with $z_{\rho_0}=0$).\vadjust{\goodbreak} The joint
probability of an observed partial ranking of length $m$ along with the
$m$ associated latent variables can be derived to be
%
\begin{eqnarray}\label{eqxandz}
&&P\bigl((X_{\rho_1},\ldots, X_{\rho_m}),(Z_1, \ldots,Z_m)|G\bigr)
\nonumber
\\
&&\qquad = P\bigl((z_{\rho_1}, \ldots, z_{\rho_m})\mbox{ and }
z_k>z_{\rho_m}\mbox{ for all }k\notin\{
\rho_1,\ldots,\rho _m\}\bigr)
\nonumber\\[-8pt]\\[-8pt]
&&\qquad = \Biggl(\prod_{i=1}^m w_{\rho_i}
\exp(-w_{\rho_i}z_{\rho_i}) \Biggr) \biggl(\prod
_{k\notin\{\rho_1,\ldots,\rho_m\}} \exp(-w_k z_{\rho
_m}) \biggr)
\nonumber
\\
&&\qquad =\prod_{i=1}^m w_{\rho_i}\exp
\Biggl(-Z_i \Biggl(\sum_{k=1}^\infty
w_k-\sum_{j=1}^{i-1}w_{\rho_j}
\Biggr) \Biggr).
\nonumber
\end{eqnarray}
Marginalising out $(Z_1, \ldots, Z_m)$ gives the probability of
$(X_{\rho_1}, \ldots, X_{\rho_m})$ as in~(\ref{eqinfinitepl}).
Further, conditional on $\rho=(\rho_i)_{i=1}^m$, it is seen that the
inter-arrival durations $Z_1, \ldots, Z_m$ are mutually independent, with\vspace*{-2pt}
\begin{eqnarray*}
Z_{i}|(X_{\rho_1}, \ldots, X_{\rho_m}),G & \sim&
\operatorname {Exp} \Biggl(\sum_{k=1}^{\infty}w_{k}-
\sum_{j=1}^{i-1}w_{\rho
_{j}} \Biggr).\vspace*{-3pt}
\end{eqnarray*}

In the next section we shall characterise the posterior distribution
over $G$ given observed partial rankings and their associated latent
variables. We end this subsection with two observations.

First, note that the jump part $\lambda(w)$ of the L\'evy intensity of
the gamma process satisfies the following property:\vspace*{-3pt}
%
\begin{eqnarray}
\int_0^{\infty} \lambda(w)\,dw&=&\infty.
\label{levyconstraint}
\end{eqnarray}
This property is equivalent (via Campbell's theorem) to the fact that
there are an infinite number of atoms in $G$ with probability one. In
other words, we are dealing with a nonparametric model with an infinite
number of choice items. It is also a necessary and sufficient condition
for the homogeneous CRM $G$ to have finite and strictly positive total
mass $0<G(\mathbb{X})<\infty$ [\citet{Regazzini2003}]. It therefore ensures that the generative
Plackett--Luce probability~(\ref{eqinfinitepl}) is well defined.


The second observation is with regard to a subtle but important
difference between the atomic measure approach described in this
section and the finite Plackett--Luce model of the previous section. In
particular, here we specified the choice items $X_k$ as locations in a
space $\mathbb{X}$ with a prior given by the base distribution $H$,
while in the finite Plackett--Luce model we simply index the $M$ choice
items using $1,\ldots,M$. One may wonder if it is possible to simply
index the infinitely many choice items using the natural numbers and
dispense with the atom locations $\{X_k\}$ altogether. This turns out
to be impossible, if we were to make the following reasonable
assumptions: that item desirabilities are a priori mutually
independent, that they are positive with probability one, and that item
desirabilities do not depend on the index of their corresponding items.
With these assumptions, along with an infinite number of choice items,
it is easy to see that the sum of all item desirabilities will be
infinite with probability one, so that the Plackett--Luce generative
model becomes ill-defined. Using the atomic measure approach, it is
possible to satisfy all assumptions while making sure the Plackett--Luce
generative model is well-defined. Note that the atoms locations $X_k$
are just used for modelling purposes. When considering inference, they
are assumed to be known and need not to be defined explicitly so as to
make inference on the item desirabilities.

\subsection{Posterior characterisation}\label{sec4.1}\label{posterior}

In this section we develop a characterisation of the posterior law of
$G$ under a gamma process prior and given Plackett--Luce observations
consisting of $L$ partial rankings. Posterior characterisation for our
model is a variation of posterior characterisation for normalised
random measures in density estimation
[\citet{Prunster2002}; \citet{James2002};
\citet{James2009}; \citet{Lijoi2010}].
We shall denote the $\ell$th partial ranking as $Y_\ell=(Y_{\ell1},
\ldots, Y_{\ell m})$, where each $Y_{\ell i}\in\mathbb{X}$. Note
that previously our partial rankings $(X_{\rho_1},\ldots, X_{\rho
_{m}})$ were denoted as ordered lists of the atoms in $G$. Since $G$ is
unobserved here, this is no longer possible, so we instead simply use a
list of observed choice items $(Y_{\ell1},\ldots, Y_{\ell m})$.
Re-expressing the conditional distribution (\ref{eqinfinitepl}) of
$Y_\ell$ given $G$, we have
\[
P(Y_\ell|G) = \prod_{i=1}^{m}
\frac{G(\{Y_{\ell i}\})}{G(\mathbb
{X}\setminus\{Y_{\ell1}, \ldots, Y_{\ell  i-1}\})}.
\]
In addition, for each $\ell$, we will also introduce a set of
auxiliary variables $Z_\ell=(Z_{\ell1}, \ldots, Z_{\ell m})$ (the
inter-arrival times) that are conditionally mutually independent given
$G$ and $Y_\ell$, with
%
\begin{eqnarray}
Z_{\ell i}|Y_\ell, G &\sim&\operatorname{Exp}\bigl(G\bigl(
\mathbb{X}\setminus \{Y_{\ell1},\ldots,Y_{\ell i-1}\}\bigr)
\bigr).\label{eqcondz}
\end{eqnarray}
The joint probability of the item lists and auxiliary variables is then
[cf. (\ref{eqxandz})]
\[
P\bigl((Y_\ell,Z_\ell)_{\ell=1}^L|G
\bigr) = \prod_{\ell=1}^L \prod
_{i=1}^{m}G\bigl(\{Y_{\ell i}\}\bigr)\exp
\bigl(-Z_{\ell i}G\bigl(\mathbb{X}\setminus\{ Y_{\ell1},
\ldots,Y_{\ell  i-1}\}\bigr)\bigr). 
\]
Note that under the generative process described in Section~\ref
{crmpl}, there is positive probability that an item appearing in a list
$Y_\ell$ appears in another list $Y_{\ell'}$ with \mbox{$\ell'\neq\ell$}.
Denote the unique items among all $L$ lists by $X^*_1, \ldots, X^*_K$,
and for each $k=1,\ldots,K$ let $n_k$ be the number of occurrences of
$X^*_k$ among the item lists. Finally, define occurrence indicators
%
\begin{eqnarray}
\delta_{\ell i k} &=& \cases{ 0, &\quad if $\exists j< i$ with
$Y_{\ell j}=X^*_k$;
\cr
1, &\quad otherwise.}\label{eqdeltas}
\end{eqnarray}
Then the joint probability under the nonparametric Plackett--Luce model is
%
\begin{eqnarray}\label{eqlikelihood}
&&P\bigl((Y_\ell,Z_\ell)_{\ell=1}^L|G
\bigr)\nonumber
\\
&&\qquad = \prod_{k=1}^K G\bigl(\bigl
\{X^*_{k}\bigr\}\bigr)^{n_k}\times \prod
_{\ell=1}^L \prod_{i=1}^{m}
\exp\bigl(-Z_{\ell i} G\bigl(\mathbb{X}\setminus\{Y_{\ell1},
\ldots,Y_{\ell  i-1}\}\bigr)\bigr)
\nonumber\\[-8pt]\\[-8pt]
&&\qquad = \exp \biggl(-G(\mathbb{X})
\sum_{\ell i}Z_{\ell i} \biggr)\nonumber
\\
&&\quad\qquad{}\times  \prod_{k=1}^K G\bigl(\bigl\{X^*_{k}\bigr\}
\bigr)^{n_k}\exp \biggl(-G\bigl(\bigl\{X^*_k\bigr\}\bigr)\sum
_{\ell
i}(\delta_{\ell ik}-1)Z_{\ell i}
\biggr). \nonumber
\end{eqnarray}
Taking expectation of (\ref{eqlikelihood}) with respect to $G$ gives
the following:

%
\begin{theorem} \label{thmmarginal}
The marginal probability of the $L$ partial rankings and latent
variables is
%
\begin{equation}\label{eqjoint}
P\bigl((Y_\ell,Z_\ell)_{\ell=1}^L\bigr)
= e^{-\psi(\sum_{\ell i}Z_{\ell i})} \prod_{k=1}^K h
\bigl(X^*_k\bigr) \kappa \biggl(n_k,\sum
_{\ell i} \delta_{\ell ik} Z_{\ell i}
\biggr),
\end{equation}
where $\psi(z)$ is the Laplace transform of $\lambda(w)$,
\[
\psi(z) = -\log\mathbb{E} \bigl[e^{-z G(\mathbb{X})} \bigr] = \int
_0^\infty\bigl(1-e^{-zw}\bigr)\lambda(w)
\,dw=\alpha\log \biggl(1+\frac{z}{\tau
} \biggr)
\]
and $\kappa(n,z)$ is the $n$th moment of the exponentially tilted
intensity $\lambda(w)e^{-zw}$:
\[
\kappa(n,z) = \int_0^\infty w^n
e^{-zw} \lambda(w)\,dw=\frac{\alpha
}{(z+\tau)^n}\Gamma(n).
\]
\end{theorem}
The proof, using the Poisson process characterisation of completely
random measures and the Palm formula [\citet{James2009}], is given in the \hyperref[sec8]{Appendix}.



Another application of the Palm formula [\citet{James2009}] now allows us to derive a posterior characterisation
of $G$. The posterior CRM can be decomposed as the sum of a CRM with
fixed atoms and a CRM whose jump part of the L\'evy intensity is
updated to $\lambda^{\ast}(w)$ in a conjugate fashion, similar to
deriving a conjugate posterior for a parametric distribution.

%
\begin{theorem} \label{thmposterior}
Given the observations and associated latent variables\break  $(Y_\ell,Z_\ell
)_{\ell=1}^L$, the posterior law of $G$ is also a gamma process, but
with atoms with both fixed and random locations. Specifically,
%
\begin{eqnarray}
G|(Y_\ell,Z_\ell)_{\ell=1}^L &=& G^* +
\sum_{k=1}^K w^*_k \delta
_{X^*_k}, \label{eqposteriorrep}
\end{eqnarray}
where $G^*$ and $w^*_1,\ldots,w^*_K$ are mutually independent. The law
of $G^*$ is still a gamma process,
\begin{eqnarray*}
G^* |(X_\ell,Z_\ell)_{\ell=1}^L &\sim&
\Gamma\bigl(\alpha,\tau^\star,H\bigr),\qquad \tau^* = \tau+\sum
_{\ell i} Z_{\ell i},
\end{eqnarray*}
while the masses have distributions,
\begin{eqnarray*}
w^*_k|(Y_\ell,Z_\ell)_{\ell=1}^L
&\sim& \operatorname{Gamma} \biggl( n_k,\tau+ \sum
_{\ell i} \delta_{\ell i k}Z_{\ell i} \biggr).
\end{eqnarray*}
\end{theorem}
\begin{pf}
Let $f\dvtx \mathbb{X}\rightarrow\mathbb{R}$ be measurable with respect
to $H$. Then
the characteristic functional of the posterior $G$ is given by
%
\begin{equation}
\qquad\mathbb{E}\bigl[e^{-\int f(x) G(dx)}|(Y_\ell,Z_\ell)_{\ell=1}^L
\bigr] = \frac{\mathbb{E}[e^{-\int f(x) G(dx)} P((Y_\ell,Z_\ell)_{\ell
=1}^L|G)]}{\mathbb{E}[P((Y_\ell,Z_\ell)_{\ell=1}^L|G)]}.\label{eqcharfunc}
\end{equation}
The denominator is as given in Theorem~\ref{thmmarginal}, while the
numerator is obtained using the same Palm formula technique as
Theorem~\ref{thmmarginal}, with the inclusion of the term $e^{-\int
f(x) G(dx)}$. Some algebra then shows that the resulting characteristic
functional of the posterior $G$ coincides with that of (\ref
{eqposteriorrep}). The proof details are given in the \hyperref[sec8]{Appendix}.
\end{pf}


\subsection{Gibbs sampling}\label{sec4.2}\label{secgibbs}
Given the results of the previous section, a simple Gibbs sampler can
now be derived, where all the conditionals are of known analytic form.
In particular, we will integrate out all of $G^*$ except for its total
mass $w^*_*=G^*(\mathbb X)$. This leaves the latent variables to
consist of the masses $w^*_*$, $(w^*_k)_{k=1}^K$ and the latent
variables $((Z_{\ell i})_{i=1}^m)_{\ell=1}^L$. The update for $Z_{\ell
i}$ is given by (\ref{eqcondz}), while those for the masses are given
in Theorem~\ref{thmposterior}:
\begin{eqnarray}\label{eqgibbst}
\mbox{Gibbs update for }Z_{\ell i}\dvt Z_{\ell i}|\operatorname{rest} &\sim&\operatorname{Exp} \biggl(w^*_*+ \sum_{k}
\delta_{\ell ik} w^*_k \biggr),\nonumber
\\
\mbox{Gibbs update for }w^*_k\dvt w^*_k|\operatorname{rest} &\sim&\operatorname{Gamma} \biggl(n_k, \tau+ \sum
_{\ell i} \delta_{\ell ik} Z_{\ell i} \biggr),
\\
\mbox{Gibbs update for }w^*_*\dvt w^*_*|\operatorname{rest} &\sim&\operatorname{Gamma}
\biggl( \alpha, \tau+ \sum_{\ell i}Z_{\ell i}
\biggr)\nonumber.
\end{eqnarray}
Note that the latent variables are conditionally independent
given the masses and vice versa.
Hyperparameters of the gamma process can be simply derived from the
joint distribution in Theorem~\ref{thmmarginal}. Since the marginal
probability of the partial rankings is invariant to rescaling of the
masses, it is sufficient to keep $\tau$ fixed at 1. As for $\alpha$,
if a $\operatorname{Gamma}(a,b)$ prior is placed on it, its
conditional distribution is still gamma:
\begin{eqnarray}
\mbox{Gibbs update for } \alpha\dvt \alpha|\operatorname{rest} &\sim&
\operatorname{Gamma} \biggl( a+K, b+\log \biggl(1+ \frac{\sum_{\ell i}Z_{\ell i}}{\tau} \biggr)
\biggr).\nonumber
\end{eqnarray}
Note that this update was derived with $w^*_*$ marginalised out, so
after an update to~$\alpha$ it is necessary to immediately update
$w^*_*$ via (\ref{eqgibbst}) before proceeding to update other variables.

In the Appendix~\ref{secgenCRM}, we show that the
construction can be extended from gamma processes to general completely
random measures, and we discuss extensions of the Gibbs sampler to this
more general case. In particular, we show that a simple Gibbs sampler
can still be derived for the generalised gamma class of completely
random measures.

\section{Mixtures of nonparametric Plackett--Luce components}\label{sec5}\label{secmixture}
In this section we propose a mixture model for heterogeneous ranking
data consisting of nonparametric Plackett--Luce components. Using the
same data augmentation scheme, we show that an efficient Gibbs sampler
can be derived and apply the model to a data set of preferences for
Irish degree programmes by high school graduates.

\subsection{Statistical model}\label{sec5.1}

Assume that we have a set of $L$ rankings $(Y_\ell)$ for $\ell\in
[L]$ of top-$m$ preferred items, and our objective is to partition
these rankings into clusters of similar preferences. We consider the
following Dirichlet process (DP) mixture model:
%
\begin{eqnarray}
\label{eqmixtdp} \pi&\sim&\operatorname{GEM}(\gamma),\nonumber
\\
c_\ell|\pi&\sim&\operatorname{Discrete}(\pi)\qquad\mbox{for }\ell = 1,
\ldots,L,
\\
Y_\ell|c_\ell,G_{c_\ell} &\sim&
\operatorname{PL}(G_{c_\ell}),
\nonumber
\end{eqnarray}
where $\operatorname{GEM}(\gamma)$ denotes the
Griffiths--Engen--McCloskey (GEM) distribution [\citet{Pitman2006}] with concentration parameter $\gamma$ (also known as the
stick-breaking construction) and $\operatorname{PL}(G)$ denotes the
nonparametric Plackett--Luce model parameterised by the atomic measure
$G$ described in Section~\ref{crmpl}. The $j$th cluster in the mixture
model is parameterised by an atomic measure $G_j$ and has mixing
proportion $\pi_j$.

To complete the model, we have to specify the prior on the component
atomic measures $G_j$. An obvious choice would be to use independent
draws from a gamma process $\Gamma(\alpha,\tau,H)$ for each $G_j$.
This unfortunately does not work. The reason is because if $H$ is
smooth, then different atomic measures will never share the same atoms.
On the other hand, notice that all items appearing in some observed
partial ranking have to come from the same Plackett--Luce model, and
thus have to appear as atoms in the corresponding atomic measure.
Putting these two observations together, the result is that any
observed pair of partial rankings that share a common item will have to
be assigned to the same component, and the mixture model will
degenerate to using a few much larger components only. In consequence,
the model will not capture the fine-scale preference structure that may
be present in the partial rankings.
This is a similar problem that motivated the hierarchical DP [\citet{TehJorBea2006a}], and the solution there, as in here, is
to allow different atomic measures to share the same set of atoms, but
to allow different atom masses.

Our solution, which is different from \citet{TehJorBea2006a}, is to make use of the Pitt--Walker [\citet{Pitt2005}] dependence model for gamma processes.
Consider a tree-structured model where there is a single root $G_0$ and
each component atomic measure $G_j$ is a leaf which connects directly
to $G_0$. The Pitt--Walker model allows us to construct the dependence
structure between the root $G_0$ and the leaves $(G_j)$ such that each
$G_j$ marginally follows a gamma process $\Gamma(\alpha,\tau,H)$. At
the root, $G_0$ is first given a gamma process prior:
\[
G_0\sim\Gamma(\alpha,\tau,H).
\]
Since $G_0$ is atomic, we can write it in the form
\[
G_0 = \sum_{k=1}^\infty
w_{0k} \delta_{X_{k}}.
\]
Now for each $j$, define a random measure $U_{j}$ with conditional law:
%
\begin{eqnarray} \label{eqhst}
U_{j}|G_0 &=& \sum_{k=1}^\infty
u_{jk} \delta_{X_{k}},
\nonumber\\[-8pt]\\[-8pt]
u_{jk}|G_0 &\sim&\operatorname{Poisson}(\phi
w_{0k}),\nonumber
\end{eqnarray}
where $\phi>0$ is a parameter which, as we shall see, governs the
strength of dependence between $G_0$ and each $G_j$. Note that since
$G_0$ has finite total mass, $U_j$~consists only of a finite number of
atoms with positive masses; the other atoms all have masses equal to zero.
Using the same Palm formula method as Section~\ref{posterior}, we can
show the following proposition:
%
\begin{proposition}\label{lempittwalker}
Suppose the prior law of $G_0$ is $\Gamma(\alpha,\tau,H)$ and $U_j$
has conditional law given by (\ref{eqhst}). The posterior law of
$G_0$ given $U_{j}$ is then
\[
G_0 = G_0^* + \sum_{k=1}^\infty
w^*_{0k} \delta_{X_{k}},
\]
where $G_0^*$ and $(w^*_{0k})_{k=1}^\infty$ are all mutually
independent. The law of $G_0^*$ is given by a gamma process
while the masses are conditionally gamma,
\begin{eqnarray*}
G_0^*|U_{j} &\sim&\Gamma(\alpha, \tau+\phi,H),
\\
w^*_{0k}|U_{j} &\sim&\operatorname{Gamma}(u_{jk},
\tau+\phi). 
\end{eqnarray*}
Note that if $u_{jk}=0$, we define $w^*_{0k}$ to be degenerate at 0,
thus, the posterior of $G_0$ consists of a finite number of atoms in
common with $U_j$, along with an infinite number of atoms (those in
$G^*_0$) not in common. The total mass of $G_0^*$ has distribution
$\operatorname{Gamma}(\alpha,\tau+\phi)$.
\end{proposition}

The idea, inspired by \citet{Pitt2005}, is to define
the conditional law of $G_{j}$ given $G_0$ and $U_{j}$ to be
independent of $G_0$ and to coincide with the conditional law of $G_0$
given $U_{j}$ as in Proposition~\ref{lempittwalker}. In other words, define
%
\begin{eqnarray}
G_{j} &=& G_{j}^* + \sum_{k=1}^\infty
w_{jk}^* \delta_{X_{k}},\label{eqgs}
\end{eqnarray}
where $G_{j}^*\sim\Gamma(\alpha,\tau+\phi,H)$ and $w_{jk}^*\sim
\operatorname{Gamma}(u_{jk},\tau+\phi)$ are mutually independent.
Note that if $u_{jk}=0$, the conditional distribution of $w_{jk}^*$
will be degenerate at 0. Hence, $G_{j}$ has an atom at $X_{k}$ if and
only if $U_{j}$ has an atom at $X_{k}$, that is, if $u_{jk}>0$. In
addition, it also has an infinite number of atoms (those in $G_{j}^*$)
which are in neither $U_{j}$ nor $G_0$.

Since the conditional laws of $G_j$ and $G_0$ given $U_j$ coincide, and
$G_0$ has prior $\Gamma(\alpha,\tau,H)$, it can be seen that $G_j$
will marginally follow the same law $\Gamma(\alpha,\tau,H)$ as well.
More compactly, we can write the dependence model as
%
\begin{eqnarray}\label{eqhierarchicalpoisgam}
U_j|G_0 &\sim&\operatorname{Poisson}(\phi G_0),
\nonumber\\[-8pt]\\[-8pt]
G_{j}|U_j &\sim&\Gamma \biggl(\alpha+U_j(
\mathbb{X}),\tau+\phi,\frac{\alpha H+U_j}{\alpha+U_j(\mathbb{X})} \biggr).
\nonumber
\end{eqnarray}

As a final observation, the parameter $\phi$ can be interpreted as
controlling the strength of dependence between $G_{0}$ and each $G_j$.
Indeed, it can be shown that
\[
\mathbb E [G_{j}|G_0] = \frac{\phi}{\phi+\tau}G_0
+ \frac{\tau
}{\phi+ \tau} H,
\]
so that larger $\phi$ corresponds to each $G_j$ being more similar to
$G_0$. \label{notephi}Larger $\phi$ may also favour a larger number
of clusters, as similar partial rankings are more likely to be
clustered in different groups.

Our construction to inducing sharing of atoms has a number of
qualitative differences from that of the hierarchical DP [\citet{TehJorBea2006a}]. First, the marginal law of each $G_j$ is
known: it is marginally a gamma process. For the hierarchical DP the
marginal laws of the individual random measures are not of simple
analytical forms. Since normalising a gamma process gives a DP, our
construction can be used as an alternative method to induce sharing of
atoms across multiple random measures, each of which still has marginal
DP law. Second, in our construction only a finite number of atoms will
be shared across random measures (though the number shared can be
controlled by the dependence parameter $\phi$), while in the
hierarchical DP all infinitely many atoms are shared.
In \citet{Caron2012a} we used the Pitt--Walker
construction for a different purpose: we
constructed a dynamical nonparametric Plackett--Luce model, where at
each time $t$, $G_t$ is a gamma process, with the Pitt--Walker
construction used to
define a Markov dependence structure for the sequence of random
measures $(G_t)$.

The structure of (\ref{eqmixtdp}), with a DP mixture with each
component specified by a random atomic measure, is reminiscent of the
nested DP of \citet{Rodriguez2008} as well, though our model has an additional
hierarchical structure allowing the sharing of atoms among different
component measures. In this respect, it also shares similarities with
the hierarchical Dirichlet process model of \citet{Muller2004}.

We focused here on a DP mixture for its simplicity, with a single
parameter $\gamma$ tuning the clustering structure. The model can be
generalised to more flexible random measures, such as Pitman--Yor
processes [\citet{Pitman1995}] or normalised random measures
[\citet{Regazzini2003}; \citet{Lijoi2007}].


%

\subsection{Posterior characterisation and Gibbs sampling}\label{sec5.2}

Assume for simplicity we have observed $L$ top-$m$ partial ranking
$Y_\ell=(Y_{\ell1},\ldots,Y_{\ell m})$ (the following will trivially
extend to partial rankings of differing sizes). We extend the results
of Section~\ref{crmpl} in characterising the posterior and developing
a Gibbs sampler for the mixture model.

Let $X^*=(X_{k}^*)_{k=1}^K$ be the set of unique items observed among
$Y_1,\ldots,Y_L$. For each cluster index $j$, let $n_{jk}$ be the
number of occurrences of item $X^*_k$ among the set of item lists
$Y_\ell$ in cluster $j$, that is, where $c_\ell=j$. Let $\rho_\ell
=(\rho_{\ell i})_{i=1}^m$ be defined such as $Y_\ell=(X^*_{\rho
_{\ell1}},\ldots,X^*_{\rho_{\ell m}})$ and $\delta_{\ell ik}$ be
occurrence indicators similar to (\ref{eqdeltas}).

\begin{figure}

\includegraphics{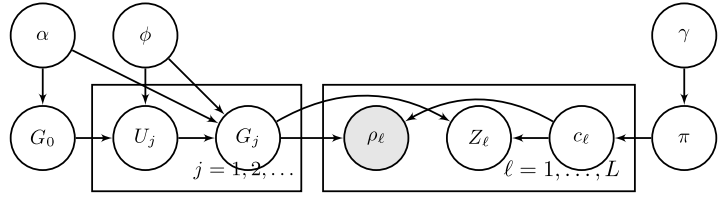}

%
\caption{Graphical model of the Dirichlet process mixture of
nonparametric Plackett--Luce components. The variables at the top are
hyperparameters, $(\rho_\ell)$ are the observed partial rankings,
while the other variables are unobserved variables.}\label{figgraphmod}%
\end{figure}

As in Section~\ref{crmpl}, the observed items $X^*$ will contain the
set of fixed atoms in the posterior law of the atomic measures $G_0, (G_j)$.
We write the masses of the fixed atoms as $w_{0k}=G_0(\{X^*_k\})$,
$w_{jk}=G_j(\{X^*_{k}\})$, while the total masses of all other random
atoms are denoted $w_{0*}=G_0(\mathbb{X}\setminus X^*)$ and
$w_{j*}=G_j(\mathbb{X}\setminus X^*)$.
We also write $u_{jk}=U_j(\{X^*_{k}\})$ and $u_{j*}=U_j(\mathbb
{X}\setminus X^*)$.
As before, we will introduce latent variables for each $\ell=1,\ldots,L$ and $i=1,\ldots,m$:
%
\begin{equation}
Z_{\ell i}|Y_\ell,c_\ell,G_{c_\ell} \sim
\operatorname{Exp} \Biggl(w_{c_\ell*} +\sum_{k=1}^K
\delta_{\ell ik} w_{c_\ell k} \Biggr). 
\label{eqlatentZt}
\end{equation}
The overall graphical model is described in
Figure~\ref{figgraphmod}.\vadjust{\goodbreak}

%


%
\begin{proposition} \label{thmposteriort}
Given the partial rankings $(Y_\ell)$ and associated latent variables
$(Z_{\ell i})$, $(u_{jk})$, $(u_{j*})$, and cluster indicators $(c_\ell
)$, the posterior law of $G_j$ is a gamma process with atoms with both
fixed and random locations. Specifically,
\begin{eqnarray*}
G_j|(Y_\ell),(Z_{\ell i}),(u_{jk}),(u_{j*}),
(c_\ell) &=& G_j^* + \sum_{k=1}^K
w_{jk} \delta_{X^*_k}, 
\end{eqnarray*}
where $G_j^*$ and $w_{j1},\ldots,w_{jK}$ are mutually independent. The
law of $G_j^*$ is a gamma process,
%
\begin{eqnarray} \label{eqGtstar}
&& G_j^* |(Y_\ell),(Z_{\ell i}),(u_{jk}),(u_{j*}),(c_\ell)
\nonumber\\[-8pt]\\[-8pt]
&&\qquad \sim \Gamma \Biggl(\alpha+u_{j*}, \tau+\phi+ \sum
_{\ell|c_\ell=j}\sum_{i=1}^m
Z_{\ell i},H \Biggr),\nonumber
\end{eqnarray}
while the masses have distributions,
%
\begin{eqnarray}\label{eqwtk}
&& w_{jk}|(Y_\ell),(Z_{\ell i}),(u_{jk}),(u_{j*}),(c_\ell)
\nonumber\\[-8pt]\\[-8pt]
&&\qquad \sim \operatorname{Gamma} \Biggl( n_{jk}+u_{jk},
\tau+\phi+ \sum_{\ell|c_\ell=j}\sum
_{i=1}^m \delta_{\ell i k}Z_{\ell i}
\Biggr)\nonumber.
\end{eqnarray}
\end{proposition}
Note that if $n_{jk}+u_{jk}=0$, then $w_{jk}=0$ and $G_j$ will not have
a fixed atom at~$X^*_k$. To complete the posterior characterisation,
note that, conditioned on $G_0$ and $G_{j}$, the variables
$u_{j1},\ldots,u_{jK}$ and $u_{j*}$ are independent, with $u_{jk}$
dependent only on $w_{0k}$ and $w_{jk}$ and similarly for $u_{j*}$. The
conditional probabilities are
%
\begin{eqnarray}
\quad p(u_{jk}|w_{0k},w_{jk})&\propto&
f_{\mathrm
{Gamma}}(w_{jk};u_{jk},\tau+\phi)f_{\mathrm
{Poisson}}(u_{jk};
\phi w_{0k}),\label{equupdate1}
\\
p(u_{j*}|w_{0*},w_{j*})&\propto&
f_{\mathrm{Gamma}}(w_{j*};\alpha +u_{j*},\tau+
\phi)f_{\mathrm{Poisson}}(u_{j*};\phi w_{0*}), \label{equupdate2}
\end{eqnarray}
where $f_{\mathrm{Gamma}}$ is the density of a Gamma
distribution and $f_{\mathrm{Poisson}}$ is the probability mass
function for a Poisson distribution.
The normalising constants are available in closed form [\citet{Mena2009}]:
%
\begin{eqnarray}
\qquad p(w_{jk}|w_{0k})&=&\exp(-\phi w_{0k})1_{w_{jk},0}
\nonumber
\\
&&{}+\mathcal I_{-1} \bigl(2\sqrt{w_{jk}\phi
w_{0k}(\tau+\phi)} \bigr) \biggl( \frac{\phi(\tau+ \phi)w_{0k}}{w_{jk}}
\biggr)^{1/2}  \label{eqcondw1}
\\
&&\quad{}\times \exp \bigl( -\phi(w_{jk}+w_{0k})-
\tau w_{jk} \bigr),\nonumber
\\
p(w_{j*}|w_{0*})&=&\mathcal I_{\alpha-1} \bigl(2
\sqrt{w_{j*}\phi w_{0*}(\tau+\phi)} \bigr) (\tau+
\phi)^{(\alpha+1)/{2}} \biggl( \frac{w_{j*}}{\phi w_{0*}} \biggr)^{(\alpha-1)/2}
\nonumber\\[-8pt] \label{eqcondw2}\\[-8pt]
&&{}\times  \exp\bigl(-\phi(w_{j*}+w_{0*})-\tau
w_{j*}\bigr),\nonumber
\end{eqnarray}
where $1_{a,b}=1$ if $a=b$, $0$ otherwise, and $\mathcal I$ is the
modified Bessel function of the first kind.
It is therefore possible to sample exactly from the discrete
distributions~(\ref{equupdate1}) and~(\ref{equupdate2}) using
standard retrospective sampling for discrete distributions; see, for example,
\citet{Papaspiliopoulos2008}.
Alternatively, we describe in the \hyperref[sec8]{Appendix} a Metropolis--Hastings
procedure that worked well in the applications.

Armed with the posterior characterisation, a Gibbs sampler can now be
derived. Each iteration of the Gibbs sampler proceeds in the following
order (details are in the \hyperref[sec8]{Appendix}):
\begin{longlist}[(2)]
\item[(1)] First note that the total masses $G_j(\mathbb X)$ are not
likelihood identifiable, so we introduce a step to improve mixing. We
simply sample them from the prior:
\begin{eqnarray*}
G_0(\mathbb X)&\sim&\operatorname{Gamma}(\alpha,\tau),
\\
U_j(\mathbb X)|G_0(\mathbb X) & \sim&
\operatorname{Poisson}\bigl(\phi G_0(\mathbb X)\bigr),
\\
G_{j}(\mathbb X)|U_j(\mathbb X) & \sim&
\operatorname{Gamma}\bigl(\alpha +U_j(\mathbb X),\tau+\phi\bigr).
\end{eqnarray*}
The individual atom masses $(w_{jk},w_{j*})$ are scaled along with the
update to the total masses.
Then the Poisson masses $(u_{jk})$, $(u_{j*})$ are updated using (\ref
{equupdate1})~and~(\ref{equupdate2}).
\item[(2)] The concentration parameter $\alpha$ and the masses $w_{0*}$,
$(w_{j*})$ and $(u_{j*})$ associated with other unobserved items are
updated efficiently using a forward--backward recursion detailed in the \hyperref[sec8]{Appendix}.
\item[(3)] The masses $(w_{0k})$ and $w_{0*}$ of the atoms in $G_0$ are
updated via an extension of Proposition~\ref{lempittwalker}. In particular,
for each item $k=1,\ldots,K$, the masses are conditionally independent
with distributions
\begin{eqnarray*}
w_{0k}|u_{1\dvtx J,k},\phi&\sim&\operatorname{Gamma} \Biggl( \sum
_{j=1}^{J}u_{jk},J
\phi+\tau \Biggr),
\end{eqnarray*}
while the total mass of the remaining atoms have conditional distribution
\begin{eqnarray*}
w_{0*}|u_{1\dvtx J*},\phi&\sim&\operatorname{Gamma} \Biggl(
\alpha+\sum_{j=1}^{J}u_{j*},J\phi+\tau \Biggr).
\end{eqnarray*}

\item[(4)] The latent variables $(Z_{\ell i})$ are updated as in (\ref
{eqlatentZt}).
\item[(5)] Conditioned on $(Z_{\ell i})$, $(u_{jk})$ and $(u_{j*})$, the
masses $(w_{jk})$ are updated via~(\ref{eqwtk}), while the total mass
of the unobserved atoms is $w_{j*}\sim\operatorname{Gamma}(\alpha
^*_j,\tau^*_j)$ from (\ref{eqGtstar}).
\item[(6)] The mixture weights $\pi$ and the allocation variables $c_\ell$
are updated using a slice sampler for mixture models [\citet{Walker2007}; \citet{Kalli2011}].
\item[(7)] Finally, the scale parameter $\gamma$ of the Dirichlet process
is updated using \citet{West1992} and the dependence parameter
$\phi$ is updated by a Metropolis--Hastings step using (\ref
{eqcondw1}) and (\ref{eqcondw2}) with the latent $(u_{jk})$ and
$(u_{j*})$ marginalised out.
\end{longlist}
The resulting algorithm is a valid partially collapsed Gibbs sampler
[\citet{VanDyk2008}]. Note, however, that permutations
of the above steps could result in an invalid sampler. The
computational cost scales as $O(K\times J\times m\times L)$, where $J$
is the average number of clusters. However, it is possible to
parallelise over the different items in the algorithm to obtain an
algorithm that scales as $O(J\times m\times L)$.

\section{Application: Irish college degree programmes}\label{sec6}
\label{seccao}

We now consider the application of the proposed model to study the
choices made by the 53,757 degree programme applicants to the College
Application Office (CAO) in the year 2000.

\subsection{Model setup and implementation details}\label{sec6.1}

\begin{figure}

\includegraphics{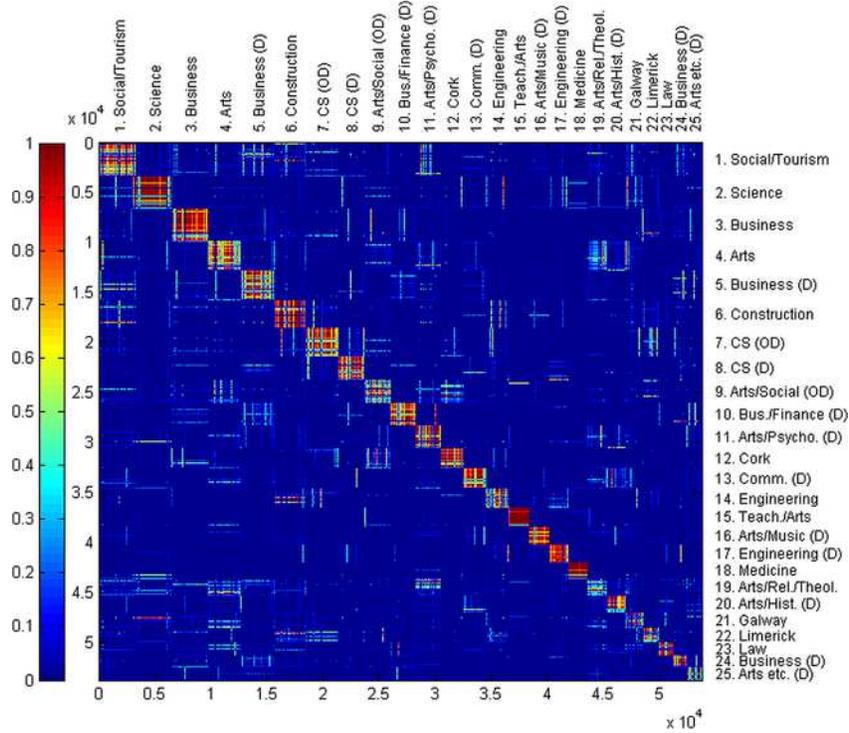}

\caption{Co-clustering matrix of the 53,757 college applicants for the
CAO data. The posterior probability that two applicants belong to the
same cluster is indicated by a color between blue (0) and red (1).
Applicants are arranged by their cluster membership, and the clusters
are ordered by size. The clusters are described in Table~\protect\ref
{tabclusters}.}\label{figcoclustering}
\end{figure}

The following flat priors are used for the hyperparameters
\begin{eqnarray*}
p(\alpha)&\propto& 1/\alpha,\qquad p(\phi)\propto1/\phi,\qquad p(\gamma)\propto 1/\gamma.
\end{eqnarray*}
We run the Gibbs sampler with $N={}$20,000 iterations. In order to obtain
a point estimate of the partition from the posterior distribution, we
use the approach proposed by \citet{Dahl2006}. Let $c^{(i)},
i=1,\ldots,N$ be the Monte Carlo samples. The point estimate $\hat c$ is obtained by
\[
\hat c = \mathop{\arg\min}_{c^{(i)}\in\{c^{(1)},\ldots,c^{(N)}\}}\sum_k
\sum_\ell(\delta_{c_k^{(i)}c_\ell^{(i)}} - \zeta
_{k\ell})^2, %
\]
where the co-clustering matrix $\zeta$ is obtained with
\[
\zeta_{k\ell}=\frac{1}{N} \sum_{i=1}^N
\delta _{c^{(i)}_kc^{(i)}_\ell} %
\]
and $\delta_{k\ell}=1$ if $k=\ell$, $0$ otherwise.
Given this partition $\hat c$, we run a Gibbs sampler with 2000
iterations to obtain the posterior mean Plackett--Luce parameters for
each cluster. Clusters are then reordered by decreasing size.
Table~\ref{tabclusters} shows the sizes of the 26 clusters which have
a size larger than 10. In addition, a co-clustering matrix was computed
based on the first MCMC run which records for each pair of students the
probability of them belonging to the same cluster. Figure~\ref
{figcoclustering} shows the co-clustering matrix to summarise the
clustering of the 53,757 students, where students are rearranged by
their cluster membership (members of the first cluster first, then
members of the second cluster, etc.).

%
\begin{table}
\tabcolsep=0pt
\caption{Description of the different clusters. The size of the
clusters, the entropy and a cluster description are provided}\label{tabclusters}
\begin{tabular*}{\tablewidth}{@{\extracolsep{\fill}}@{}lccc@{}}
\hline
\textbf{Cluster} & \textbf{Size} & \textbf{Entropy} & \textbf{Description}\\
\hline
\phantom{0}1 & 3325 & 0.72 & Social science/tourism \\
\phantom{0}2 & 3214 & 0.71 & Science                  \\
\phantom{0}3 & 3183 & 0.64 & Business/commerce          \\
\phantom{0}4 & 2994 & 0.58 & Arts                         \\
\phantom{0}5 & 2910 & 0.63 & Business/marketing-Dublin    \\
\phantom{0}6 & 2879 & 0.68 & Construction                 \\
\phantom{0}7 & 2803 & 0.66 & CS-outside Dublin            \\
\phantom{0}8 & 2225 & 0.67 & CS-Dublin                    \\
\phantom{0}9 & 2303 & 0.67 & Arts/social-outside Dublin   \\
10 & 2263 & 0.63 & Business/finance-Dublin     \\
11 & 2198 & 0.65 & Arts/psychology-Dublin      \\
12 & 2086 & 0.63 & Cork                        \\
13 & 2029 & 0.64 & Comm./journalism-Dublin
\\
14 & 1918 & 0.71 & Engineering\\
15 & 1835 & 0.48 & Teaching/arts \\
16 & 1835 & 0.68 & Art/music-Dublin \\
17 & 1740 & 0.71 & Engineering-Dublin \\
18 & 1701 & 0.55 & Medicine \\
19 & 1675 & 0.70 & Arts/religion/theology \\
20 & 1631 & 0.76 & Arts/history-Dublin\\
21 & 1627 & 0.66 & Galway \\
22 & 1392 & 0.70 &Limerick \\
23 & 1273 & 0.65 & Law \\
24 & 1269 & 0.72 &Business-Dublin \\
25 & 1225 & 0.79 & Arts/bus./language-Dublin \\
26 & \phantom{00}47 & 0.96 & Mixed \\
\hline
\end{tabular*}
\end{table}

%

\subsection{Results}\label{sec6.2}

An examination of the Plackett--Luce parameter for each cluster reveals
that the subject matter of the degree programme is a strong determinant
of the clustering of students (Table~\ref{tabclusters}). For example,
clusters 6, 18 and 23 are characterised as construction, medicine and
law, respectively.
Besides the type of degree, geographical location is a strong
determinant of degree programme choice. Clusters 12, 21 and 22 are,
respectively, concerned with applications to college degree programmes
in Cork, Galway and Limerick. There is a lot of heterogeneity in the
subject area of the college degree programmes for these clusters, as
can be seen, for example, for the Cork cluster 12 in Table~\ref
{tabcork}. A number of clusters are also defined by a combination of
both subject area and location, for example, for clusters 7 and 8 in
Tables~\ref{tabCSnotDublin} and~\ref{tabCSDublin}, which correspond
to computer science, respectively, outside and inside Dublin.

%
\begin{table}
\tabcolsep=0pt
\caption{Cluster 12: Cork}\label{tabcork}
\begin{tabular*}{\tablewidth}{@{\extracolsep{\fill}}@{}lccc@{}}
\hline
\textbf{Rank} & \textbf{Aver. norm. weight} & \textbf{College} & \textbf{Degree programme} \\
\hline
\phantom{0}1 & 0.105 & University College Cork & Arts \\
\phantom{0}2 & 0.072 & University College Cork & Computer science \\
\phantom{0}3 & 0.072 & University College Cork & Commerce \\
\phantom{0}4 & 0.067 & University College Cork & Business information systems \\
\phantom{0}5 & 0.057 & Cork IT & Computer applications \\
\phantom{0}6 & 0.049 & Cork IT & Software dev. and comp. net. \\
\phantom{0}7 & 0.035 & University College Cork & Finance \\
\phantom{0}8 & 0.031 & University College Cork & Law \\
\phantom{0}9 & 0.031 & University College Cork & Accounting \\
10 & 0.026 & University College Cork & Biological and chemical sciences
\\
\hline
\end{tabular*}
\end{table}
%

%
\begin{table}
\tabcolsep=0pt
\caption{Cluster 7: Computer science-outside Dublin}\label{tabCSnotDublin}
\begin{tabular*}{\tablewidth}{@{\extracolsep{\fill}}@{}lccc@{}}
\hline
\textbf{Rank} & \textbf{Aver. norm. weight} & \textbf{College} & \textbf{Degree programme} \\
\hline
\phantom{0}1 & 0.081 & Cork IT & Computer applications \\
\phantom{0}2 & 0.075 & Limerick IT & Software development \\
\phantom{0}3 & 0.072 & University of Limerick & Computer systems \\
\phantom{0}4 & 0.064 & Waterford IT & Applied computing \\
\phantom{0}5 & 0.061 & Cork IT & Software dev. and  comp. net. \\
\phantom{0}6 & 0.046 & IT Carlow & Computer networking \\
\phantom{0}7 & 0.038 & Athlone IT & Computer and software engineering \\
\phantom{0}8 & 0.036 & University College Cork & Computer science \\
\phantom{0}9 & 0.033 & Dublin City University & Computer applications \\
10 & 0.033 & University of Limerick & Information technology \\
 \hline
\end{tabular*}
\end{table}
%

%
\begin{table}
\tabcolsep=0pt
\caption{Cluster 8: Computer science-Dublin}\label{tabCSDublin}
\begin{tabular*}{\tablewidth}{@{\extracolsep{\fill}}@{}lccc@{}}
\hline
\textbf{Rank} & \textbf{Aver. norm. weight} & \textbf{College} & \textbf{Degree programme} \\
\hline
\phantom{0}1 & 0.141 & Dublin City University & Computer applications \\
\phantom{0}2 & 0.054 & University College Dublin & Computer science \\
\phantom{0}3 & 0.049 & NUI-Maynooth & Computer science \\
\phantom{0}4 & 0.043 & Dublin IT & Computer science \\
\phantom{0}5 & 0.040 & National College of Ireland & Software systems \\
\phantom{0}6 & 0.038 & Dublin IT & Business info. systems dev. \\
\phantom{0}7 & 0.036 & Trinity College Dublin & Computer science \\
\phantom{0}8 & 0.035 & Dublin IT & Applied sciences/computing \\
\phantom{0}9 & 0.030 & Trinity College Dublin & Information and comm. tech. \\
10 & 0.029 & University College Dublin & B.A. (computer science)\\
\hline
\end{tabular*}
\end{table}

As mentioned in Section~\ref{collegeapplications}, there is a common
perception in the Irish society and media that students pick degree
programme based on prestige rather than subject area. Another
perception is that the points requirement for a degree programme is a
measure of prestige; in fact, the points requirement is determined by a
number of factors including the number of available places, the number
of applicants who list the degree programme in their top-10 preferences
and the quality of the applicants who apply for the degree programme.
Such a selection-by-prestige phenomenon should be evidenced by a
cluster of students picking degree programmes in medicine and law, both
of which have very high points requirements, but no such cluster was
found. In fact, medicine and law applicants are clustered separately
into clusters 18 and 23, respectively. Therefore, the clustering
suggests that students are primarily picking degree programmes on the
basis of subject area and geographical considerations; this finding is
in agreement with the results found in
\citet{Gormley2006}; \citet{McNicholas2007}.

It is also of interest to look at the variability of the student
choices within each cluster. This can be quantified by the normalised
entropy, which takes its values between 0 and 1, and defined for each
cluster $j$ by
\[
\frac{-\sum_{k=1}^K  (\hat w_{jk} \log\hat w_{jk}
) - \hat w_{j*} \log\hat w_{j*}}{\log(K+1)},
\]
where $\hat w_{jk}$ are the averaged normalised weights of item $k$
in cluster $j$ obtained from the second MCMC run; the normalised
entropy values for each cluster are reported in Table~\ref
{tabclusters}. A low value indicates low variability in the choices
within a cluster, whereas a large value indicates a lot of variability.
Interestingly, cluster 15 has very low normalised entropy, where 56\%
of the students in that cluster are likely to take one of the three
most popular degree programmes of that cluster (Drumcondra, Froebel or
Marina) as their first choice; these degree programmes are the main
primary teacher education degree programmes in Dublin and, thus, many
members of this cluster have a strong interest in teacher education as
a degree choice. Further, there is much more variability in cluster 7,
where students choices are spread across various computing degree
programmes, and only 23\% of the students are likely to take one of the
three most popular degree programmes as their first choice.

The co-clustering matrix reveals some interesting connections between
clusters, which have not been explored in previous analyses of the CAO
data. For example, the plot reveals that a number of applicants have
high probability of belonging to clusters 4 and 19 which are both in
the arts. Cluster 4 is characterised by arts degrees which do not
require the applicants to select their major in advance, whereas
cluster 19 is characterised by arts degrees where the student needs to
specify their major in advance. It is worth observing that the clusters
are fairly well separated, and very few clusters exhibit the phenomenon
of sharing applicants, which is further evidence that the applicants
are only selecting degree programmes of a particular type (as described
by the cluster names in Table~\ref{tabclusters}).

Marginal Posterior distributions of the hyperparameters $\alpha$,
$\gamma$ and $\phi$ are, respectively, in the ranges $[3,8]$, $[2,5]$
and $[100,200]$. Correlation parameter $\phi$ is rather high. This is
due to the fact that some degree programmes, such as Arts in University
College Dublin or Cork, often appear in the top-ten list of applicants,
whatever their main subject matter is. Parameter $\gamma$ is
associated to the number of clusters, which is around 35. Parameter
$\alpha$ relates to the variability of the weights within clusters
(and thus to the entropy of the clusters).

\section{Discussion}\label{sec7}\label{secdiscussion}

We have proposed a Bayesian nonparametric Plackett--Luce model for
ranked data. Our approach is based on the theory of completely random
measures, where we showed that the Plackett--Luce generative model
corresponds exactly to a size-biased permutation of the atoms in the
random measure. We characterised the posterior distribution and derived
a simple MCMC sampling algorithm for posterior simulation. Our approach
can be seen as a multi-stage generalisation of posterior inference in
normalised random measures [\citet{Regazzini2003};
\citet{James2009}; \citet{Griffin2011}; \citet{Favaro2013}].




We also developed a nonparametric mixture model consisting of
nonparametric Plackett--Luce components to model heterogeneity in
partial ranking data. In order to allow atoms to be shared across
components, we made use of the Pitt--Walker construction, which was
previously only used to define Markov dynamical models. Applying our
model to a data set of preferences for Irish college degree programmes,
we find interesting clustering structure supporting the observation
that students were choosing programmes mainly based on subject area and
geographical considerations.

It is worthwhile comparing our mixture model to another nonparametric
mixture model, DPM-GM, where each component is a generalised Mallows
model [\citet{Busse2007}; \citet{Meila2008};
\citet{Meila2010}]. In the
generalised Mallows model the component distributions are characterised
by a (discrete) permutation parameter, whereas in the Plackett--Luce
model the component distributions are characterised by a continuous
rating parameter. Thus, the Plackett--Luce model offers greater
modelling flexibility to capture the strength of preferences for each
item. On the other hand, the scale parameters in the generalised
Mallows model can accommodate varying precision in the ranking.
\mbox{Additionally,} inference for the generalised Mallows models can be difficult.

The mixture model established the existence of clusters of applicants
with similar degree programme preferences and characterises these
clusters and their coherence in terms of choices. The results support
the previous hypotheses that subject matter and geographical location
are the primary drivers of degree programme choice [\citet{Gormley2006};
\citet{McNicholas2007}]. These factors
are important because they reflect the intrinsic interest in the
subject matter of the degree programmes and the economic and practical
aspects of choosing a third level institution for study. The
geographical location influence is further supported by results on
acceptances to degree programmes [\citet{OConnell2006}] and studies on how students fund their education which
found that 45\% of Irish university students live in their family home
[\citet{Clancy1999}] and thus attend an institution
that is geographically close by.

An interesting extension of the proposed model would be to consider
inhomogeneous completely random measures, where the preferences would
depend on a set of covariates (e.g., location).

\begin{appendix}
\section{Proof of Theorem~\texorpdfstring{\lowercase{\protect\ref{thmmarginal}}}{1}}\label{sec8}
The marginal probability (\ref{eqjoint}) is obtained by taking the
expectation of (\ref{eqlikelihood}) with respect to $G$.
Note however that (\ref{eqlikelihood}) is a density, so to be totally
precise here we need to work with the \emph{probability} of
infinitesimal neighborhoods around the observations instead, which
introduces significant notational complexity. To keep the notation
simple, we will work with densities, leaving it to the careful reader
to verify that the calculations indeed carry over to the case of probabilities.
%
%
\begin{eqnarray*}
&& P\bigl((Y_\ell,Z_\ell)_{\ell=1}^L
\bigr)
\nonumber
\\
&&\qquad = \mathbb{E} \bigl[P\bigl((Y_\ell,Z_\ell)_{\ell=1}^L|G
\bigr) \bigr]
\nonumber
\\
&&\qquad = \mathbb{E} \Biggl[ e^{-G(\mathbb{X})\sum_{\ell i}Z_{\ell i}} \prod_{k=1}^K
G\bigl(\bigl\{X^*_{k}\bigr\}\bigr)^{n_k} e^{-G(\{X^*_k\})\sum_{\ell i}(\delta_{\ell ik}-1)Z_{\ell i}}
\Biggr].
\nonumber
\end{eqnarray*}
The gamma prior on $G=\sum_{j=1}^\infty w_j\delta_{X_j}$ is
equivalent to a Poisson process prior on $N=\sum_{j=1}^\infty\delta
_{(w_j,X_j)}$ defined over the space $\mathbb{R}^+\times\mathbb{X}$
with mean intensity $\lambda(w)h(x)$. Then,
\begin{eqnarray}
\qquad &&= \mathbb{E} \Biggl[ e^{-\int w N(dw,dx) \sum_{\ell i}Z_{\ell i}} \prod_{k=1}^K
\sum_{j=1}^\infty w_j^{n_k}
\mathbh{1}\bigl(X_j=X^*_{k}
\bigr)e^{-w_j\sum
_{\ell i}(\delta_{\ell ik}-1)Z_{\ell i}} \Biggr].
\end{eqnarray}
We now recall the Palm formula [see e.g., \citet{Bertoin2006},
Lemma~2.3].

%
\begin{proposition}
Palm Formula. Let $N$ be a Poisson process on $S$ with mean measure
$\nu$.
Let $S_{p}$ denote the set of point measures on $S$, $f\dvtx S\rightarrow
{}[0,+\infty{}[$ and $\mathcal G\dvtx S\times S_{p}\rightarrow
{}[0,+\infty
{}[$ be some measurable functional. Then we have the so-called Palm
formula%
%
\begin{equation}\label{eqPalmFormula}
\mathbb{E} \biggl[ \int_{S}f(x)\mathcal G(x,N)N(dx)
\biggr] =\int_{S}\mathbb{E}%
\bigl[\mathcal
G(x,N+dx)\bigr]f(x)\nu(dx),
\end{equation}
where the expectation is with respect to $N$.
\end{proposition}
Applying the Palm formula for Poisson processes to pull the $k=1$ term
out of the expectation,
%
\begin{eqnarray}
&&= \int\mathbb{E} \Biggl[ e^{-\int w (N+\delta_{w^*_1,x^*_1})(dw,dx)
\sum_{\ell i}Z_{\ell i}} \prod_{k=2}^K
\sum_{j=1}^\infty w_j^{n_k}
\mathbh{1}\bigl(X_j=X^*_{k}
\bigr) e^{-w_j\sum_{\ell i}(\delta_{\ell ik}-1)Z_{\ell i}} \Biggr]
\nonumber
\\
&&\hspace*{20pt}{} \times \bigl(w_1^*\bigr)^{n_1}h
\bigl(X^*_1\bigr) e^{-w_1^*\sum_{\ell i}(\delta_{\ell
i1}-1)Z_{\ell i}}\lambda\bigl(w^*_1
\bigr)\,dw^*_1
\nonumber
\\
&&= \mathbb{E} \Biggl[ e^{-\int w N(dw,dx) \sum_{\ell i}Z_{\ell i}} \prod_{k=2}^K
\sum_{j=1}^\infty w_j^{n_k}
\mathbh{1}\bigl(X_j=X^*_{k}
\bigr) e^{-w_j\sum_{\ell i}(\delta_{\ell ik}-1)Z_{\ell i}} \Biggr]
\nonumber
\\
&&\hspace*{9pt}{} \times h\bigl(X^*_1\bigr) \int\bigl(w_1^*
\bigr)^{n_1} e^{-w_1^*\sum_{\ell i}\delta_{\ell
i1}Z_{\ell i}}\lambda\bigl(w^*_1
\bigr)\,dw^*_1.
\nonumber
\end{eqnarray}
Now iteratively pull out terms $k=2,\ldots,K$ using the same idea, and
we get:
\begin{eqnarray}\label{eqdenominator}
&& = \mathbb{E} \bigl[ e^{-G(\mathbb{X}) \sum_{\ell i}Z_{\ell i}} \bigr] \prod_{k=1}^Kh
\bigl(X^*_k\bigr)\int\bigl(w_k^*\bigr)^{n_k}
e^{-w_k^*\sum_{\ell i}\delta
_{\ell ik}Z_{\ell i}}\lambda\bigl(w^*_k\bigr)\,dw^*_k
\nonumber\\[-8pt]\\[-8pt]
&& = e^{-\psi (\sum_{\ell i}Z_{\ell i} )} \prod_{k=1}^K h
\bigl(X^*_k\bigr) \kappa \biggl(n_k,\sum
_{\ell i}\delta_{\ell
ik}Z_{\ell i}
\biggr).\nonumber
\end{eqnarray}
This completes the proof of Theorem~\ref{thmmarginal}.

\section{Proof of Theorem~\texorpdfstring{\lowercase{\protect\ref{thmposterior}}}{2}}\label{sec9}

The proof is essentially obtained by calculating the numerator and
denominator of (\ref{eqcharfunc}). The denominator is already given
in Theorem~\ref{thmmarginal}. The numerator is obtained using the
same technique with the inclusion of the term $e^{\int f(x) G(dx)}$,
which gives
%
\begin{eqnarray}
&&\mathbb{E} \bigl[e^{-\int f(x) G(dx)} P\bigl((Y_\ell,Z_\ell)_{\ell
=1}^L|G
\bigr) \bigr]
\nonumber
\\
&&\qquad = \mathbb{E} \bigl[e^{-\int(f(x)+\sum_{\ell i}Z_{\ell i})
G(dx)} \bigr] \nonumber
\\
&&\quad\qquad{}\times \prod_{k=1}^K
h\bigl(X^*_k\bigr) \int \bigl(w_k^*\bigr)^{n_k}
e^{-w_k^*(f(X^*_k)+\sum_{\ell i}\delta_{\ell ik}Z_{\ell
i})}\lambda\bigl(w^*_k\bigr)\,dw^*_k.
\nonumber
\end{eqnarray}
By the L\'evy--Khintchine theorem (using the fact that $G$ has a Poisson
process representation $N$),
\begin{eqnarray}\label{eqnumerator}
&= &\exp \biggl(-\int\bigl(1-e^{-w(f(x)+\sum_{\ell i}Z_{\ell i})}\bigr) \lambda (w)h(x)\,dw\,dx \biggr)
\nonumber\\[-8pt]\\[-8pt]
&&{}\times \prod_{k=1}^K h
\bigl(X^*_k\bigr) \int \bigl(w_k^*\bigr)^{n_k}
e^{-w_k^*(f(X^*_k)+\sum_{\ell i}\delta_{\ell ik}Z_{\ell
i})}\lambda\bigl(w^*_k\bigr)\,dw^*_k.\nonumber
\end{eqnarray}
Dividing the numerator (\ref{eqnumerator}) by the denominator (\ref
{eqdenominator}), the characteristic functional of the posterior $G$ is
\begin{eqnarray*}
&&\mathbb{E} \bigl[e^{-\int f(x)G(dx)}|(Y_\ell,Z_\ell)_{\ell
=1}^L
\bigr]
\nonumber
\\
&&\qquad =\exp \biggl(-\int\bigl(1-e^{-wf(x)}\bigr)e^{-\sum_{\ell i}Z_{\ell i}} \lambda
(w)h(x)\,dw\,dx \biggr)
\nonumber
\\
&&\quad\qquad{}\times \prod_{k=1}^K h
\bigl(X^*_k\bigr) \frac{
\int e^{-f(X^*_k)} (w_k^*)^{n_k} e^{-w_k^*\sum_{\ell i}\delta_{\ell
ik}Z_{\ell i}}\lambda(w^*_k)\,dw^*_k
}{
\int(w_k^*)^{n_k} e^{-w_k^*\sum_{\ell i}\delta_{\ell ik}Z_{\ell
i}}\lambda(w^*_k)\,dw^*_k
}.
\end{eqnarray*}
Since the characteristic functional is the product of $K+1$ terms, we
see that the posterior $G$ consists of $K+1$ independent components,
one corresponding to the first term above ($G^*$), and the others
corresponding to the $K$ terms in the product over $k$.
Substituting the L\'evy measure $\lambda(w)$ for a gamma process, we
note that the first term shows that $G^*$ is a gamma process with
updated inverse scale $\tau^*$. The $k$th term in the product shows
that the corresponding component is an atom located at $X^*_k$ with
density $(w_k^*)^{n_k} e^{-w_k^*\sum_{\ell i}\delta_{\ell ik}Z_{\ell
i}}\lambda(w^*_k)$; this is the density of the gamma distribution over
$w^*_k$ in Theorem~\ref{thmposterior}. This completes the proof.

\section{Generalisation to completely random~measures}\label{sec10}
\label{secgenCRM}
The posterior characterisation we have developed along with the Gibbs
sampler can be easily extended to completely random measures (CRM)
[\citet{Kin1967a}; \citet{Regazzini2003}; \citet{Lijoi2010}]. To keep the exposition simple, we
shall consider homogeneous CRMs without fixed atoms. These can be
described, as for the gamma process before, with atom locations $\{X_k\}
$ i.i.d. according to a nonatomic base distribution $H$, and with atom
masses $\{w_k\}$ being distributed according to a Poisson process over
$\mathbb{R}^+$ with a general L\'evy measure $\lambda(w)$ which
satisfies the constraints (\ref{levyconstraint}) leading to a
normalisable measure $G$ with infinitely many atoms. We will write
$G\sim\operatorname{CRM}(\lambda,H)$ if $G$ follows the law of a
homogeneous CRM with L\'evy intensity $\lambda(w)$ and base
distribution $H$.

Both Theorems~\ref{thmmarginal} and \ref{thmposterior} generalise
naturally to homogeneous CRMs. In fact the statements and the proofs in
the appendix still hold with the more general L\'evy intensity, along
with its Laplace transform $\psi(z)$ and moment function $\kappa(n,z)$:

{\renewcommand{\thetheoremstar}{\arabic{theoremstar}$'$}
%
\begin{theoremstar} \label{margstar}
The marginal probability of the $L$ partial rankings and latent
variables is
\[
P\bigl((Y_\ell,Z_\ell)_{\ell=1}^L\bigr)
= e^{-\psi(\sum_{\ell i}Z_{\ell i})} \prod_{k=1}^K h
\bigl(X^*_k\bigr) \kappa \biggl(n_k,\sum
_{\ell i} \delta_{\ell ik} Z_{\ell i}
\biggr), 
\]
where $\psi(z)$ is the Laplace transform of $\lambda(w)$,
\[
\psi(z) = -\log\mathbb{E} \bigl[e^{-z G(\mathbb{X})} \bigr] = \int
_0^\infty\bigl(1-e^{-zw}\bigr)\lambda(w)
\,dw
\]
and $\kappa(n,z)$ is the $n$th moment of the exponentially tilted L\'
evy intensity $\lambda(w)e^{-zw}$:
\[
\kappa(n,z) = \int_0^\infty w^n
e^{-zw} \lambda(w)\,dw.
\]
\end{theoremstar}

%
\begin{theoremstar} \label{poststar}
Given the observations and associated latent variables $(Y_\ell,Z_\ell
)_{\ell=1}^L$, the posterior law of $G$ is also a homogeneous CRM, but
with atoms with both fixed and random locations. Specifically,
\begin{eqnarray*}
G|(Y_\ell,Z_\ell)_{\ell=1}^L &=& G^* +
\sum_{k=1}^K w^*_k \delta
_{X^*_k}, 
\end{eqnarray*}
where $G^*$ and $w^*_1,\ldots,w^*_K$ are mutually independent. The law
of $G^*$ is a homogeneous CRM with an exponentially tilted L\'evy intensity:
\begin{eqnarray*}
G^* |(X_\ell,Z_\ell)_{\ell=1}^L &\sim&
\operatorname{CRM}\bigl(\lambda ^\star,H\bigr),\qquad \lambda^*(w) =
\lambda(w)e^{-w\sum_{\ell i} Z_{\ell i}}
\end{eqnarray*}
while the masses have densities:
\begin{eqnarray*}
P\bigl(w^*_k|(Y_\ell,Z_\ell)_{\ell=1}^L
\bigr) &=& \frac
{(w^*_k)^{n_k}e^{-w^*_k\sum_{\ell i} Z_{\ell i}}\lambda
(w^*_k)}{\kappa(n_k,\sum_{\ell i} Z_{\ell i})}.
\end{eqnarray*}
\end{theoremstar}}%

Examples of CRMs that have been explored in the literature for Bayesian
nonparametric modelling include the stable process
[\citet{Kin1975a}], the inverse Gaussian process
[\citet{LijMenPru2005a}], the generalised gamma process [\citet{Bri1999a}],
and the beta process [\citet{Hjo1990a}]. The generalised gamma process
forms the largest known simple and tractable family of CRMs, with the
gamma, stable and inverse Gaussian processes included as subfamilies.
It has a L\'evy intensity of the form
\[
\lambda(w) = \frac{\alpha}{\Gamma(1-\sigma)}w^{-1-\sigma}e^{-\tau w},
\]
where the concentration parameter is $\alpha>0$, the inverse scale is
$\tau\ge0$, and the index is $0\le\sigma<1$. The gamma process is
recovered when $\sigma=0$, the stable when $\tau=0$, and the inverse
Gaussian when $\sigma=1/2$. The Laplace transform and the moment
function of the generalised gamma process are
\begin{eqnarray*}
\psi(z) &=& \frac{\alpha}{\sigma}\bigl((\tau+z)^{\sigma}-\tau^{\sigma}
\bigr),\qquad \kappa(n,z) = \frac{\alpha}{(\tau+z)^{n-\sigma}}\frac{\Gamma
(n-\sigma)}{\Gamma(1-\sigma)}.
\end{eqnarray*}\eject\noindent

The Gibbs sampler developed for the gamma process can be generalised to
homogeneous CRMs as well. Recall that given the observed partial
rankings, the parameters consist of the ratings $(w^*_k)_{k=1}^K$ of
the observed items and the total ratings $w^*_*$ of the unobserved
ones, while the latent variables are $(Z_{\ell i})$. A corollary of
Theorems \ref{margstar} and \ref{poststar} which will prove useful is
the joint probability of these along with the observed partial rankings:
%
\begin{eqnarray}\label{eqjointstar2}
&& P\bigl((Y_{\ell i},Z_{\ell i}),\bigl(w^*_k\bigr),w^*_*
\bigr)
\nonumber\\[-8pt]\\[-8pt]
&&\qquad  = e^{-w^*_*(\sum_{\ell i} Z_{\ell i})} f\bigl(w^*_*\bigr) \prod_{k=1}^K
h\bigl(X^*_k\bigr) \bigl(w^*_k\bigr)^{n_k}e^{-w^*_k(\sum_{\ell i}\delta_{\ell ik} Z_{\ell i})}
\lambda\bigl(w^*_k\bigr),\nonumber
\end{eqnarray}
where $f(w)$ is the density (assumed to exist) of the total mass
$w^*_*$ under a CRM with the \emph{prior} L\'evy intensity $\lambda(w)$.
Note that integrating out the parameters $(w^*_k),w^*_*$ from (\ref
{eqjointstar2}) gives the marginal probability in Theorem \ref
{margstar}. From the joint\vadjust{\goodbreak} probability (\ref{eqjointstar2}), the
Gibbs sampler can now be derived:
\begin{eqnarray*}
\mbox{Gibbs update for }Z_{\ell i}\dvt Z_{\ell i}|\operatorname{rest}
&\sim&\operatorname{Exp} \biggl(w^*_*+ \sum_{k}\delta_{\ell ik} w^*_k \biggr),
\\
\mbox{Gibbs update for }w^*_k\dvt P\bigl(w^*_k|\operatorname{rest}\bigr) &\propto&\bigl(w^*_k\bigr)^{n_k}e^{-w^*_k\sum_{\ell i}
Z_{\ell i}}
\lambda\bigl(w^*_k\bigr),
\\
\mbox{Gibbs update for }w^*_*\dvt P\bigl(w^*_*|\mbox{rest}\bigr) &\propto&
e^{-w^*_*(\sum_{\ell i} Z_{\ell i})} f\bigl(w^*_*\bigr).
\end{eqnarray*}
To be concrete, consider the updates for a generalised gamma process.
The conditional distribution for $w^*_k$ can be seen to be
$\operatorname{Gamma}(n_k-\sigma,\tau+\sum_{\ell i} Z_{\ell i})$,
while the conditional distribution for $w^*_*$ can be seen to be an
exponentially tilted stable distribution. This is not a standard
distribution (nor does it have known analytic forms for its density),
but can be effectively sampled using recent techniques [\citet{Dev2009a}]. Another approach is to marginalise out $w^*_*$ first:
\[
P\bigl((Y_{\ell i},Z_{\ell i}),\bigl(w^*_k\bigr)\bigr) =
e^{-\psi(\sum_{\ell i} Z_{\ell i})}\prod_{k=1}^K h
\bigl(X^*_k\bigr) \bigl(w^*_k\bigr)^{n_k}e^{-w^*_k(\sum_{\ell i}\delta_{\ell ik} Z_{\ell i})}
\lambda\bigl(w^*_k\bigr). 
\]
The MCMC algorithm then consists of sampling the ratings $(w^*_k)$ and
auxiliary variables $(Z_{\ell i})$.
Marginalising out $w^*_*$ introduces additional dependencies among the
latent variables $Z_{\ell i}$. Fortunately, since the Laplace transform
for a generalised gamma process is of simple form, it is possible to
update the latent variables $(Z_{\ell i})$ using a variety of standard
techniques, including Metropolis--Hastings, Hamiltonian Monte Carlo, or
adaptive rejection sampling. For these techniques to work well we
suggest reparametrising each $Z_{\ell i}$ using its logarithm $\log
Z_{\ell i}$ instead.

\section{Gibbs sampler for the mixture of nonparametric Plackett--Luce components}\label{sec11}
\label{secgibbsmixturePL}

Let $J$ be the number of different values taken by $c$ (number of
clusters). Please note that the number of clusters is not set in
advance and its value may change at each iteration.
The Gibbs sampler proceeds with each of the following updates in turn:
\begin{enumerate}[3.]
\item[1.]
\begin{enumerate}[(a)]
\item[(a)] Update $G_0(\mathbb{X})$ given $\alpha$, then for
$j=1,\ldots,J$, update $G_j(\mathbb{X})$ given $(G_0(\mathbb
{X}),\alpha,\phi,c)$.

\item[(b)] For $j=1,\ldots,J$, update $(u_{j},u_{j*})$ given
($w_{0},w_{0*},w_{j},w_{j*},\phi,\alpha,c)$.
\end{enumerate}
\item[2.]
\begin{enumerate}[(a)]
\item[(a)] Update $\alpha$ given $(Z,\phi,c)$.

\item[(b)] Update $w_{0*}$ given $(Z,\phi,c,\alpha)$.

\item[(c)] For $j=1,\ldots,J$, update $u_{j*}$ given $(Z,\phi,c,\alpha,w_{0*})$.

\item[(d)] For $j=1,\ldots,J$, update $w_{j*}$ given $ ( Z,\alpha,u_{j*},\phi,c ) $.
\end{enumerate}
\item[3.] Update $(w_{0k}),w_{0*}$ given $(U_{1\dvtx J},\alpha)$.

\item[4.] For $\ell=1,\ldots,L$, update $Z_\ell$ given $(w_{c_\ell
},w_{c_\ell*},c_\ell)$.

\item[5.] For $j=1,\ldots,J$, update $(w_{j},w_{j*})$ given $ (
Z,\alpha,u_{j},u_{j*},\phi,c ) $.

\item[6.] For $\ell=1,\ldots,L$, update $c_\ell$ and the mixture
weights $\pi$ given $w_{1\dvtx J},w_{1\dvtx J*}$.

\item[7.] Update $\gamma$ given $c$.

\item[8.] Update $\phi$ given $w_{0},w_{0*},w_{1\dvtx J},w_{1\dvtx J*},\alpha,\phi$.
\end{enumerate}
%
%

The step are now fully described.

%
\begin{longlist}
\item[1(a)] \textit{Update $G_0(\mathbb{X})$ given $\alpha$}, \textit{then for $j=1,\ldots,J$}, \textit{update} $G_j(\mathbb{X})$ \textit{given} $(G_0(\mathbb
{X}),\alpha,\phi,c)$

We have
\[
G_{0}(\mathbb{X})|\alpha\sim\operatorname{Gamma}(\alpha,\tau)
\]
and for $j=1,\ldots,J$
\[
G_{j}(\mathbb{X})\sim\operatorname{Gamma}(\alpha+M_{j},
\tau+\phi),
\]
where $M_{j}\sim\operatorname{Poisson}(\phi G_{0}(\mathbb{X}))$.
\end{longlist}
\begin{longlist}
\item[1(b)] \textit{For $j=1,\ldots,J$}, \textit{update} $(u_{j},u_{j*})$ \textit{given}
($w_{0},w_{0*},w_{j},w_{j*},\phi,\alpha,c)$

Consider first the sampling of $u_{j}$. We have, for $j=1,\ldots,J$ and
$k=1,\ldots,K$
\[
p(u_{jk}|w_{0k},w_{jk})\propto
p(u_{jk}|w_{0k}%
)p(w_{jk}|u_{jk}),
\]
where
\[
p(u_{jk}|w_{0k})=f_{\mathrm{Poisson}}(u_{jk};\phi
w_{0k})
\]
and
\[
p(w_{jk}|u_{jk})=\cases{ \delta_{0}(w_{jk}),
&\quad if $u_{jk}=0$,
\cr
f_{\mathrm{Gamma}}(w_{jk};u_{jk},
\tau+\phi), &\quad if $u_{jk}>0$.}
\]
Hence we can have the following MH update. If $w_{jk}>0$, then we
necessarily have $u_{jk}>0$. We sample $u_{jk}^{\ast}\sim
$zPoisson$(\phi
w_{0k})$ where zPoisson$(\phi w_{0k})$ denotes the zero-truncated
Poisson distribution and accept $u_{jk}^{\ast}$ with probability
\[
\min \biggl( 1,\frac{f_{\mathrm{Gamma}}(w_{jk};u_{jk}^{\ast
},\tau+\phi
)}{f_{\mathrm{Gamma}}(w_{jk};u_{jk},\tau+\phi)} \biggr).
\]

If $w_{jk}=0$, we only have two possible moves: $u_{jk}=0$ or $u_{jk}%
=1$, given by the following probabilities
\begin{eqnarray*}
P(u_{jk}  = 0|w_{jk}=0,w_{0k})&=&
\frac{\exp(-\phi w_{0k}%
)}{\exp(-\phi w_{0k})+\phi w_{0k}\exp(-\phi w_{0k})(\tau+\phi
)}
\\
&=&\frac{1}{1+\phi w_{0k}(\tau+\phi)},
\\
P(u_{jk}  = 1|w_{jk}=0,w_{0k})&=&
\frac{\phi w_{0k}\exp
(-\phi w_{0k})(\tau+\phi)}{\exp(-\phi w_{0k})+\phi w_{0k}%
\exp(-\phi w_{0k})(\tau+\phi)}
\\
&=&\frac{\phi w_{0k}(\tau+\phi
)}{1+\phi w_{0k}(\tau+\phi)}.
\end{eqnarray*}

Note that the above Markov chain is not irreducible, as the probability is
zero to go from a state $ ( u_{jk}>0,w_{jk}>0 ) $ to a state
$ ( u_{jk}=0,w_{jk}=0 ) $, even though the posterior
probability of this event is nonzero in the case item $k$ does not
appear in
cluster $j$. We can add such moves by jointly sampling $(u_{jk},w_{jk}%
)$. For each $k$ that does not appear in cluster $j$, sample
$u_{jk}^{\ast
}\sim\operatorname{Poisson}(\phi w_{0k})$ then set $w_{jk}^{\ast}=0$ if
$u_{jk}^{\ast}=0$ otherwise sample $w_{jk}^{\ast}\sim\operatorname
{Gamma}%
(u_{jk},\tau+\phi)$. Accept $(u_{jk}^{\ast},w_{jk}^{\ast})$ with
probability
\[
\min \biggl( 1,\frac{\exp(-w_{jk}^{\ast}\sum_{\ell|c_{\ell
}=j}\sum_{i=1}%
^{m}Z_{\ell i})}{\exp(-w_{jk}\sum_{\ell|c_{\ell}=j}\sum_{i=1}^{m}Z_{\ell i}%
)} \biggr).
\]

We now consider sampling of $u_{j*}$, $j=1,\ldots,J$. We can use a MH
step. Sample $w_{j*}^{\ast}\sim\operatorname{Poisson}(\phi w_{0*})$
and accept with
probability
\[
\min \biggl( 1,\frac{f_{\mathrm{Gamma}}(u_{j*};\alpha
+u_{j*}^{\ast},\tau+\phi)}%
{f_{\mathrm{Gamma}}(u_{j*};\alpha+u_{j*}^{\ast},\tau+\phi
)} \biggr).
\]
\end{longlist}
\begin{longlist}
\item[2(a)] \textit{Update $\alpha$ given $(Z,\phi,c)$}

We can sample from the full conditional which is given by
\[
\alpha|(Z,\gamma,\phi,c)\sim\operatorname{Gamma} \bigl( a+K,b+y_{0}+
\log(1+x_{0}%
) \bigr),
\]
where
\begin{eqnarray*}
x_{0} & =& \sum_{j=1}^{J}
\frac{\phi\widetilde Z_{j}}{1+\phi
+\widetilde Z_{j}},
\\
y_{0} & =& -\sum_{j=1}^{J}\log
\biggl( \frac{1+\phi}{1+\phi
+\widetilde Z_{j}} \biggr)
\end{eqnarray*}
with $\widetilde Z_{j}=\sum_{\ell|c_{\ell}=j}\sum_{i=1}^{m}Z_{\ell i}$.
\end{longlist}
\begin{longlist}
\item[2(b)] \textit{Update} $w_{0*}$ \textit{given} $(Z,\phi,c,\alpha)$

We can sample from the full conditional which is given by
\[
w_{0*}|(Z,\phi,c,\alpha)\sim\operatorname{Gamma} ( \alpha,
\tau+x_0 ),
\]
where $x_0$ is defined above.
\end{longlist}
\begin{longlist}
\item[2(c)] \textit{For} $j=1,\ldots,J$, \textit{update} $u_{j*}$ \textit{given}
$(Z,\phi,c,\alpha,w_{0*})$

We can sample from the full conditional which is given, for $j=1,\ldots,J$ by
\[
u_{j*}|(Z,\phi,c,\alpha,w_{0*})\sim\operatorname{Poisson}
\biggl(\frac{1+\phi}{1+\phi+\widetilde Z_j}\phi w_{0*} \biggr),
\]
where $\widetilde Z_j$ is defined above.
\end{longlist}
\begin{longlist}
\item[2(d)] \textit{For} $j=1,\ldots,J$, \textit{update} $w_{j*}$ \textit{given}
$( Z,\alpha,u_{j*},\phi,c )$

We can sample from the full conditional which is given, for $j=1,\ldots,J$ by
\[
w_{j*}|u_{j*},Z,c,\alpha\sim\operatorname{Gamma} ( \alpha
+u_{j*},\tau+\phi+\widetilde Z_j ),
\]
where $\widetilde Z_j$ is defined above.
\end{longlist}
\begin{longlist}
\item[3.] \textit{Update $(w_{0k}),w_{0*}$ given $(U_{1\dvtx J},\alpha)$}

For each item $k=1,\ldots,K$, sample
\[
w_{0k}|u_{1\dvtx J,k},\phi\sim\operatorname{Gamma} \Biggl( \sum
_{j=1}^{J}u_{jk},J
\phi+\tau \Biggr).
\]

Sample the remaining mass
\[
w_{0*}|u_{1\dvtx J*},\phi\sim\operatorname{Gamma} \Biggl( \alpha+
\sum_{j=1}^{J}u_{j*},J\phi+\tau \Biggr).
\]
\end{longlist}
\begin{longlist}
\item[4.] \textit{For} $\ell=1,\ldots,L$, \textit{update} $Z_\ell$ \textit{given}
$(w_{c_\ell},w_{c_\ell*},c_\ell)$

For $\ell=1,\ldots,L$ and $i=1,\ldots, m$, sample
\[
Z_{\ell i}|c,w,w_*\sim\operatorname{Exp} \Biggl( w_{c_{\ell},*} +\sum
_{k=1}^{K}\delta_{\ell ik}
w_{c_{\ell},k}%
\Biggr). 
\]
\end{longlist}
\begin{longlist}
\item[5.] \textit{For} $j=1,\ldots,J$, \textit{update} $(w_{jk}),w_{j*}$
\textit{given} $ ( Z,\alpha,u_{j},u_{j*},\phi,c ) $

For each cluster $j=1,\ldots,J$

\begin{itemize}
\item For each item $k=1,\ldots,K$, sample
\[
w_{jk}|u_{jk},\{\rho_{\ell}|c_{\ell}=j\}
\sim\operatorname {Gamma} \Biggl( n_{jk}%
+u_{jk},
\tau+\phi+\sum_{\ell|c_{\ell}=j} \Biggl\{ \sum
_{i=1}^{m}\delta _{\ell ik}Z_{\ell i}
\Biggr\} \Biggr)
\]
if $u_{jk}+n_{jk}>0$, otherwise, set $w_{jk}=0$.

\item Sample the total mass
\[
w_{j*}|u_{j*},\{\rho_{\ell}|c_{\ell}=j\}
\sim\operatorname {Gamma} \Biggl( \alpha+u_{j*},\tau+\phi+
\sum_{\ell|c_{\ell}=j}\sum_{i=1}^{m}Z_{\ell i}
\Biggr).
\]
\end{itemize}
\end{longlist}
\begin{longlist}
\item[6.] \textit{For} $\ell=1,\ldots,L$, \textit{update} $c_\ell$ \textit{and the
weights} $\pi$ \textit{given} $w_{1\dvtx J},w_{1\dvtx J*}$

The allocation variables $(c_{1},\ldots,c_L)$ are updated using the
slice sampling technique described in [\citet{Walker2007};
\citet{Kalli2011}; \citet{Fall2012}].\label{secslicesampler} It builds on
the \hyperref[sec1]{Introduction} of additional latent slice variables, and does not
require to set any truncation. For completeness, we briefly recall here
the details of the sampler. From equation~(\ref{eqmixtdp}), we have
%
\begin{equation}
f(Y_\ell|\pi,G)= \sum_{k=1}^\infty
\pi_k PL(Y_\ell;G_k),
\end{equation}
where the $\pi_k$ admit the following stick-breaking representation
%
\begin{equation}
\pi_1 =v_1, \pi_k=v_k\prod
_{j<k}(1-v_j),
\end{equation}
where the $v_k$ are i.i.d. from $\operatorname{Beta}(1,\gamma)$. For
each observation $Y_\ell$, slice sampling introduces latent variable
$\omega_\ell$ such that the joint distribution of $Y_\ell$, $\omega
_\ell$ and $c_\ell$ is given by
%
\begin{equation}
f(Y_\ell,\omega_\ell,c_\ell|\pi,G)= 1(
\omega_\ell<\pi_{c_\ell
})\operatorname{PL}(Y_\ell;G_{c_\ell}).
\end{equation}

For simplicity, assume that the $c_\ell$ take values in $\{1,2,\ldots,J\}$. Let $\mu_k$ be the number of allocation variables taking value
$k\in\{1,\ldots,J\}$. The sampler samples $\omega$ and $v$ as a
block given $c$, then $c$ given $v$ and $\omega$. 
%
\begin{enumerate}[2.]
\item[1.]
\begin{enumerate}[(a)]
\item[(a)] Sample $(\pi_1,\ldots,\pi_J,\pi_*)\sim\operatorname
{Dirichlet}(\mu_1,\ldots,\mu_J,\gamma)$.

\item[(b)] For $\ell=1,\ldots,L$, sample $\omega_\ell\sim\operatorname
{Unif}([0,\pi_{c_\ell}])$.

\item[(c)] Set $k=J$. While $\sum_{j=1}^k \pi_k< (1-\min(\omega_1,\ldots,\omega_L))$.
\begin{itemize}
\item Set $k=k+1$.
\item Sample $v_k\sim\operatorname{Beta}(1,\gamma)$.
\item Set $\pi_k = \pi_* v_k \prod_{j=J+1}^{k-1}(1-v_j)$.
\item Sample $G_k$ given $G_0$ using equation (\ref{eqhierarchicalpoisgam}).
\end{itemize}
\end{enumerate}

\item[2.] For $\ell=1,\ldots,L$, sample $c_\ell$ from
\[
p(c_\ell=k)\propto1(\pi_k>\omega_\ell)
\operatorname{PL}(Y_\ell;G_{c_\ell}).
\]
\end{enumerate}
\end{longlist}
\begin{longlist}
\item[7.] \textit{Update} $\gamma$ \textit{given} $c$

The scale parameter $\gamma$ of the Dirichlet process is updated using
the data augmentation technique of \citet{West1992}.
\end{longlist}
\begin{longlist}
\item[8.] \textit{Update} $\phi$ \textit{given}
$w_{0},w_{0*},w_{1\dvtx J},w_{1\dvtx J*},\alpha,\phi$

We sample $\phi$ using a MH step. Propose $\phi^{\ast}=\phi\exp
(\sigma
\varepsilon)$ where $\sigma>0$ and $\varepsilon\sim\mathcal
N(0,1)$. And accept it with
probability
\[
\min \Biggl( 1,\frac{p(\phi^{\ast})}{p(\phi)}\frac{\phi^{\ast
}}{\phi}%
\prod
_{j=1}^{J} \Biggl[\frac{p(w_{j*}|\phi^{\ast
},w_{0*})}{p(w_{j*}|\phi,w_{0*})}%
\prod
_{k=1}^{K}\frac{p(w_{jk}|\phi^{\ast},w_{0k})}{p(w
_{jk}|\phi,w_{0k})} \Biggr] \Biggr).
\]
\end{longlist}
\end{appendix}

\section*{Acknowledgments}
The authors thank Igor Pr\"unster for very helpful feedback on an
earlier version of this work. Fran\c cois Caron acknowledges the support of the
European Commission under the Marie Curie Intra-European Fellowship
Programme.\footnote{The contents reflect only the authors views and
not the views of the European Commission.}



%

\printaddresses

\end{document}